\documentclass[10pt,twocolumn,letterpaper]{article}

\usepackage[pagenumbers]{cvpr} 

\usepackage{graphicx}
\graphicspath{{images/}{./}}  
\usepackage{booktabs}
\usepackage{multirow}
\usepackage{amsmath,amssymb}
\usepackage{xcolor}
\usepackage{enumitem}
\usepackage{makecell}
\usepackage{xspace}
\usepackage{microtype}     
\usepackage[most]{tcolorbox}      
\usepackage{algorithm}            
\usepackage{algpseudocode}
\usepackage{tabularx}             
\usepackage{rotating}             
\usepackage{listings}             
\usepackage{longtable}            
\usepackage{needspace}            
\usepackage{placeins}             

\newcommand{\TODO}[1]{}

\newcommand{\bench}{PInVerify\xspace}
\newcommand{\task}{Active Instance Verification\xspace}
\newcommand{\taskShort}{AIV\xspace}

\DeclareMathOperator*{\argmax}{arg\,max}

\renewcommand{\paragraph}[1]{\vspace{.4em}\noindent\textbf{#1.}}

\definecolor{cvprblue}{rgb}{0.21,0.49,0.74}
\usepackage[pagebackref,breaklinks,colorlinks,allcolors=cvprblue]{hyperref}

\def\paperID{14}
\def\confName{CVPR}
\def\confYear{2026}

\title{PInVerify: An Offline Embodied Benchmark for Active Instance Verification}

\author{Yuhang Jiang\\
University of Trento, Italy\\
{\tt\small jyhtjtj@gmail.com}}

\begin{document}
\maketitle

\begin{abstract}
Embodied agents have made strong progress in navigating to target objects, but reaching the goal vicinity does not guarantee that the agent has found the \emph{correct} instance: subtle attribute differences (e.g., ``white floral'' vs.\ ``white striped'') often require close-range, multi-view inspection. We address this gap with \task{} (\taskShort{}), a task in which an agent actively selects viewpoints around a candidate object to decide whether it matches a fine-grained natural-language description. We formalize \taskShort{} as a finite-horizon decision process and introduce \bench{}, an offline embodied benchmark for \taskShort{}: 3{,}000 evaluation episodes across 18 object categories, delivered as multi-view captures with a 6-sector navigation topology that exposes \emph{trap views} (navigable but uninformative) and \emph{unreachable sectors}. As reference baselines we build a training-free pipeline and a LoRA-fine-tuned end-to-end agent around open-source multimodal large language models (MLLMs) at on-device scale ($\leq$8B parameters), with attribute decomposition, a visibility-weighted multi-view tracker, and three next-best-view (NBV) strategies. In our evaluation across Qwen3-VL (4B/8B), SenseNova-SI-1.2-InternVL3-8B, CLIP, and SigLIP2, the best MLLM-based baseline exceeds the best embedding baseline by 4.9\,pp; GT-box ablations show a +3.1\,pp detection gap; and we do not observe reliable gains from active viewpoint selection within the tested NBV strategies. A LoRA-fine-tuned agent (SFT+GSPO) reaches 85.6\%. \bench{} aims to support further work on active, fine-grained semantic verification in embodied AI. Code: \url{https://github.com/Avalon-S/PInVerify}.
\end{abstract}

\section{Introduction}
\label{sec:intro}

Embodied AI has moved from category-level navigation toward fine-grained instance understanding. ObjectNav benchmarks on Matterport3D/HM3D scenes~\cite{chang2017matterport3d,ramakrishnan2021habitat,chaplot2020object} ask agents to find an object category, HM3D-OVON~\cite{yokoyama2024hm3d} expands the goal vocabulary to free-form language, and PIN~\cite{barsellotti2024personalized} moves to user-specific instances described by language. Yet these tasks still rely on arrival- or proximity-oriented success: the episode terminates at a goal vicinity rather than asking for an explicit pre-interaction verification decision over the perceived object.

This assumption masks a real failure mode. Given ``fetch the blue mug with white floral patterns,'' an agent that reaches a blue mug with white stripes still scores as a hit on an arrival-style navigation metric, even though deployment would treat it as a wrong delivery. RGB sensors at navigation distance may fail to resolve fine-grained patterns, and discriminative attributes can be distributed across non-frontal surfaces. Recent designs such as CompassNav~\cite{li2025compassnav} insert a ``Target Verification'' step into the agent's reasoning chain, but verification remains embedded inside navigation rather than benchmarked as a standalone decision problem.

\begin{figure*}[t]
  \centering
  \includegraphics[width=\linewidth]{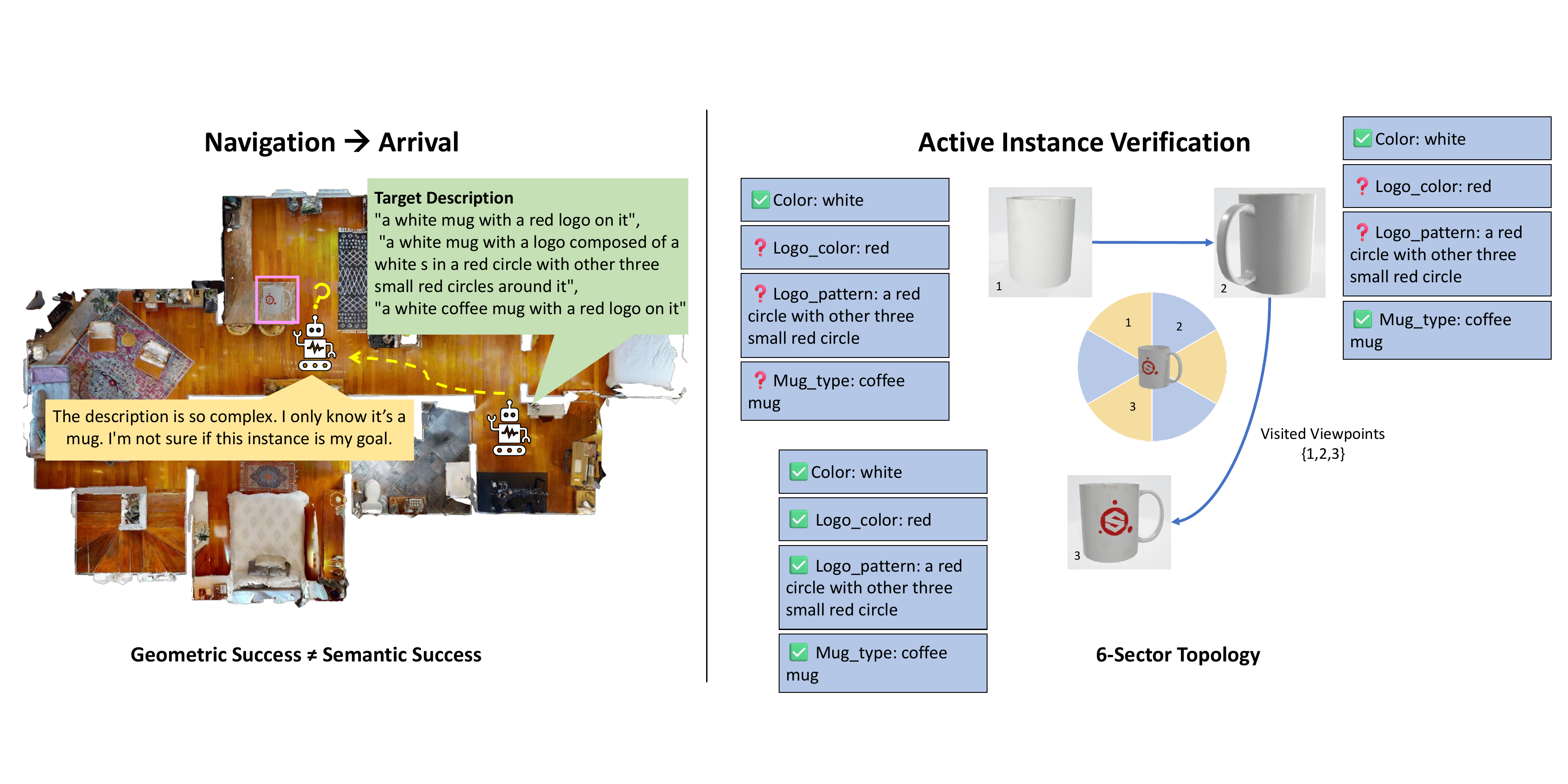}
  \caption{\textbf{Geometric success vs.\ semantic success.} Left: arrival-based navigation declares success at goal vicinity, but the agent has not actually verified instance identity. Right: \taskShort{} addresses this gap by selecting multiple viewpoints around the candidate and progressively confirming attributes until a confident decision is reached.}
  \label{fig:overview}
\end{figure*}

We argue that an active verification stage before interaction is needed to close this semantic gap, and we formalize it as \task{} (\taskShort{}). Figure~\ref{fig:overview} contrasts arrival-based navigation with \taskShort{}: given a candidate object and a fine-grained description, the agent selects a sequence of viewpoints and decides whether the object matches.

\taskShort{} differs from related work along three axes. Navigation benchmarks (ObjectNav~\cite{chaplot2020object}, OVON~\cite{yokoyama2024hm3d}, PIN~\cite{barsellotti2024personalized}) ask how to reach a target; \taskShort{} begins after arrival and asks whether the target is correct. Active 3D reconstruction~\cite{dunn2009next, vasquez2014reconstruction, jin2023neu} optimizes geometric completeness; \taskShort{} optimizes attribute-level semantic confidence. RefCOCO and its variants~\cite{yu2016modeling, mao2016generation} evaluate fixed 2D observations; \taskShort{} operates under the occlusion and viewpoint uncertainty inherent to embodied perception.

\paragraph{Contributions}
\begin{enumerate}[leftmargin=1.4em,itemsep=2pt]
  \item \textbf{Task.} We define \task{} (\taskShort{}) as a finite-horizon decision process: an agent verifies whether a candidate object matches a fine-grained natural-language description by actively selecting viewpoints around it (\S\ref{sec:task}).
  \item \textbf{Benchmark.} We introduce \bench{}, an offline embodied benchmark for \taskShort{}: 3{,}000 evaluation episodes across 18 categories, realized as multi-view captures over a 6-sector navigation graph, with protocol-level annotations (trap views, unreachable sectors, stratified positive / neg\_same / neg\_diff pairs) absent from upstream navigation datasets (\S\ref{sec:benchmark}).
  \item \textbf{Reference agents and findings.} We provide training-free and LoRA-fine-tuned reference agents and use them to map out the difficulty of \bench{}: the best tested MLLM baseline is stronger than the best embedding baseline in our evaluation, GT-box ablations identify detection quality as a major bottleneck, our tested open-source $\leq$8B MLLMs do not show reliable gains from the active action space, and the best query mode appears to depend on the model's calibration profile, not just its raw capability (\S\ref{sec:agent}--\S\ref{sec:experiments}).
\end{enumerate}
\noindent\bench{} is meant as a controlled testbed: by abstracting locomotion away from the verification decision, it complements navigation benchmarks rather than replacing them.

\section{Related Work}
\label{sec:related}

\paragraph{Embodied benchmarks for navigation}
Mainstream embodied benchmarks evaluate physical interaction or navigation rather than pre-interaction semantic verification. BEHAVIOR-1K~\cite{li2023behavior}, iGibson~2.0~\cite{li2021igibson}, and ARNOLD~\cite{gong2023arnold} target manipulation and long-horizon planning. Within navigation, ObjectNav benchmarks on Matterport3D/HM3D scenes~\cite{chang2017matterport3d,ramakrishnan2021habitat,chaplot2020object} require reaching any object of a goal category, while HM3D-OVON~\cite{yokoyama2024hm3d} and GOAT-Bench~\cite{khanna2024goat} extend navigation to open-vocabulary and lifelong settings. Building on the language-instructed navigation paradigm of R2R~\cite{anderson2018vision}, PIN~\cite{barsellotti2024personalized} and REVERIE~\cite{qi2020reverie} push toward language-specified instance targets. Across this lineage, metrics focus on navigation or grounding at the stopping point; we are not aware of a benchmark in this group that isolates pre-interaction instance verification as the primary task.

\paragraph{Active vision and next-best-view}
Classical next-best-view (NBV) research targets geometric completeness via structure-from-motion or active 3D reconstruction~\cite{dunn2009next, vasquez2014reconstruction}. Related learned 3D reconstruction work ranges from dense scene reconstruction from posed RGB images~\cite{murez2020atlas} to generalizable RL policies for 3D coverage~\cite{chen2024gennbv} and uncertainty-driven neural rendering/mapping for view planning~\cite{jin2023neu, xue2024neural}. A semantic line shifts NBV toward discovery and recognition: goal-oriented semantic exploration for ObjectNav~\cite{chaplot2020object} and uncertainty-guided open-vocabulary navigation~\cite{bajpai2025uncertainty} use semantic uncertainty to find new targets, while evidential active recognition~\cite{fan2024evidential} sequentially accumulates per-view evidence with explicit uncertainty quantification for open-world category recognition. \taskShort{} is scrutiny-driven rather than discovery-driven: it takes an already-found candidate and selects views to reduce attribute-level ambiguity rather than spatial uncertainty or category identity. Closest in spirit on the recognition side, \citet{du2023multiview} pose multi-view active fine-grained recognition with policy-gradient view selection on a vehicle dataset; \taskShort{} differs in that the task is verification against a free-form natural-language description rather than $N$-way classification, queries are language-conditioned, and the setting is an embodied 3D scene with trap views and unreachable sectors. Concurrent MLLM-based active perception has been trained via GRPO for 2D image grounding~\cite{zhu2025active} and as a learned active view-selection policy that feeds refined views to a VLM verifier for visual question answering~\cite{koo2025ambulatory}; \taskShort{} differs in targeting language-conditioned instance verification on a discrete sector graph with multi-view belief accumulation rather than 2D zoom-in or single-view refinement.

\paragraph{Referring expressions and embodied QA}
Referring expression comprehension on RefCOCO and its variants~\cite{yu2016modeling, mao2016generation} evaluates linguistic--visual grounding in static 2D images. Embodied question answering spans agents that actively navigate to gather evidence~\cite{das2018embodied} and OpenEQA's episodic-memory and active-exploration settings~\cite{majumdar2024openeqa}. Static embodied benchmarks~\cite{xiao2025static} evaluate pre-recorded scene observations, while spatial/perceptual QA benchmarks (VSI-Bench~\cite{yang2024thinking}, EmbSpatial-Bench~\cite{du2024embspatial}, 3DSRBench~\cite{ma20253dsrbench}, BLINK~\cite{fu2024blink}) probe reasoning over supplied imagery or video rather than requiring the agent to choose new views in the environment. \taskShort{} extends the referring expression paradigm to embodied 3D contexts where the agent acquires new evidence under occlusion and viewpoint uncertainty before committing.

Two adjacent methods recognize this verification gap within end-to-end navigation: IEVE~\cite{lei2024instance} switches among Exploration, Verification, and Exploitation actions during image-goal navigation, and CompassNav~\cite{li2025compassnav} inserts a ``Target Verification'' step into prompted reasoning. We instead isolate verification as a standalone evaluation stage with language-conditioned queries (vs.\ IEVE's image goals) and benchmark-level supports (vs.\ CompassNav's prompt-level verification step). Table~\ref{tab:bench_compare} (\S\ref{sec:bench_compare}) positions \bench{} relative to related benchmarks.

\section{The \taskShort{} Task}
\label{sec:task}

We formalize \task{} (\taskShort{}) as a finite-horizon decision process. An episode is a tuple $(\mathcal{Q}, o_{\text{query}}, o_{\text{target}}, \mathcal{G})$, where $\mathcal{Q} = \{q_1, q_2, q_3\}$ is a set of natural-language descriptions characterizing the \emph{query instance} $o_{\text{query}}$, $o_{\text{target}}$ is the candidate object physically placed in the scene, and $\mathcal{G} = (\mathcal{V}, \mathcal{E})$ is a navigation graph around $o_{\text{target}}$ whose nodes are pre-captured viewpoints and whose edges represent navigable transitions between sectors. The agent must determine whether $o_{\text{target}} = o_{\text{query}}$ from $\mathcal{Q}$ and the sequence of observations it actively chooses.

\paragraph{State, action, and horizon}
At step $t$, the state is $S_t = (p_t, I_t, B_t)$, where $p_t \in \mathcal{V}$ is the agent's current viewpoint, $I_t$ is the observation (the RGB image at $p_t$, plus a visibility warning when the previous \texttt{NAV} action landed on a trap view or unreachable sector; see \S\ref{sec:trap}), and $B_t$ is the agent's internal belief about the verification decision so far. We deliberately leave $B_t$'s form open: an attribute-level evidence record (one realization in \S\ref{sec:agent}), an accumulated reasoning trace, or a learned latent. Candidate localization within $I_t$ is delegated to a pluggable detector module: Grounding DINO~\cite{liu2024grounding} by default, ground-truth bounding boxes from the capture metadata for ablation (\S\ref{sec:experiments}). \taskShort{} thus primarily evaluates verification, with detection quality as an explicit evaluation axis. The action space is
\begin{equation}
  \mathcal{A} = \{\texttt{NAV}_{d_1}, \ldots, \texttt{NAV}_{d_K}\} \cup \{\texttt{YES}, \texttt{NO}\},
\end{equation}
with $K = 6$ azimuthal sectors; \texttt{YES}/\texttt{NO} are terminal verification decisions and \texttt{NAV}$_{d_k}$ moves to the sector in relative direction $d_k$. The horizon is $T = 6$ (each sector visited at most once), transitions on the offline graph $\mathcal{G}$ are deterministic, and episodes initialize from a sector where the target is visible (\texttt{valid\_start\_sectors}, App.~\ref{app:index_generation}).

\paragraph{Pair types, labels, and objective}
Each episode is a verification pair $(o_{\text{query}}, o_{\text{target}})$ with label $y = 1$ if $o_{\text{query}} = o_{\text{target}}$ and $y = 0$ otherwise. \bench{} further stratifies negatives into two difficulty levels:
\begin{itemize}[leftmargin=1.4em,itemsep=2pt]
  \item \textbf{neg\_same}: same category, different fine-grained attributes (e.g., two backpacks differing in color or pattern). Tests instance-level discrimination.
  \item \textbf{neg\_diff}: different categories (e.g., backpack query vs.\ hat target). Acts as a category-level sanity check.
\end{itemize}
The evaluation set is balanced 1{:}1{:}1 across the three pair types, separating calibration bias (over-confirming vs.\ over-rejecting) from category-level recognition and supporting the failure-mode analysis in \S\ref{sec:experiments}. The agent maximizes verification accuracy while minimizing observation steps; we report accuracy and Average Steps to Decision (ASD) as the two primary metrics, with additional diagnostics in \S\ref{sec:metrics}.

\section{The \bench{} Benchmark}
\label{sec:benchmark}

\bench{} provides 3{,}000 evaluation episodes for \taskShort{}. We describe its data source (\S\ref{sec:data_source}), the multi-view capture pipeline that produces the offline navigation graph (\S\ref{sec:capture}), the trap-view and unreachable-sector annotations that distinguish \bench{} from a static photo collection (\S\ref{sec:trap}), the evaluation protocol (\S\ref{sec:metrics}), and a comparison with related benchmarks (\S\ref{sec:bench_compare}).

\subsection{Data source: building on PInNED}
\label{sec:data_source}

\bench{} uses the Personalized Instance Navigation with Elaborate Descriptions (PInNED) dataset~\cite{barsellotti2024personalized} as the substrate for scene geometry, object placements, and language descriptions. PInNED provides photorealistic indoor scenes from HM3D~\cite{ramakrishnan2021habitat} rendered via the Habitat simulator~\cite{savva2019habitat}, with 338 Objaverse-XL~\cite{deitke2023objaverse} object instances injected across 18 everyday categories\footnote{Backpack, bag, ball, book, camera, cellphone, eyeglasses, hat, headphones, keys, laptop, mug, shoes, teddy bear, toy, visor, wallet, watch.} and three natural-language descriptions per instance. Of the 338 instances, 335 are usable; three are excluded due to renderer-level texture anomalies or post-placement visibility failures (the object lands in a position where no navigable viewpoint admits a valid mask).

Like ObjectNav on MatterPort3D, \bench{} reuses an existing visual substrate but defines a different task and ships new annotations on top of it. Table~\ref{tab:pinverify_vs_pinned} lists what is reused versus newly added.

\begin{table}[t]
\centering
\caption{What is new in \bench{} relative to PInNED.}
\label{tab:pinverify_vs_pinned}
\setlength{\tabcolsep}{3pt}
\small
\begin{tabular}{@{}l c c@{}}
\toprule
\textbf{Element} & \textbf{PInNED} & \textbf{\bench{}} \\
\midrule
\multicolumn{3}{@{}l}{\textit{Reused}} \\
\quad HM3D scenes & \checkmark & \checkmark \\
\quad Object placements & \checkmark & \checkmark \\
\quad Per-instance descriptions & \checkmark & \checkmark \\
\midrule
\multicolumn{3}{@{}l}{\textit{New in \bench{}}} \\
\quad Task: nav-arrival $\to$ verification & --- & \checkmark \\
\quad 6-sector capture topology & --- & \checkmark \\
\quad Far / near observation rings & --- & \checkmark \\
\quad Per-capture segmentation mask & --- & \checkmark \\
\quad Trap-view annotation & --- & \checkmark \\
\quad Sector-level reachability metadata & --- & \checkmark \\
\quad Stratified eval pairs\textsuperscript{$\dagger$} & --- & \checkmark \\
\quad Offline navigation graph $\mathcal{G}$ & --- & \checkmark \\
\bottomrule
\end{tabular}\\[2pt]
\raggedright\footnotesize $\dagger$ PIN injects same- and different-category distractor objects into the scene. \bench{} reuses these distractor pools but routes them differently: distractors provide the \emph{query description} while the original target stays in the scene, yielding a controlled 1{:}1{:}1 stratification at the evaluation-protocol level rather than at the scene level.
\end{table}

\subsection{Multi-view capture pipeline}
\label{sec:capture}

\begin{figure*}[t]
  \centering
  \includegraphics[width=\linewidth]{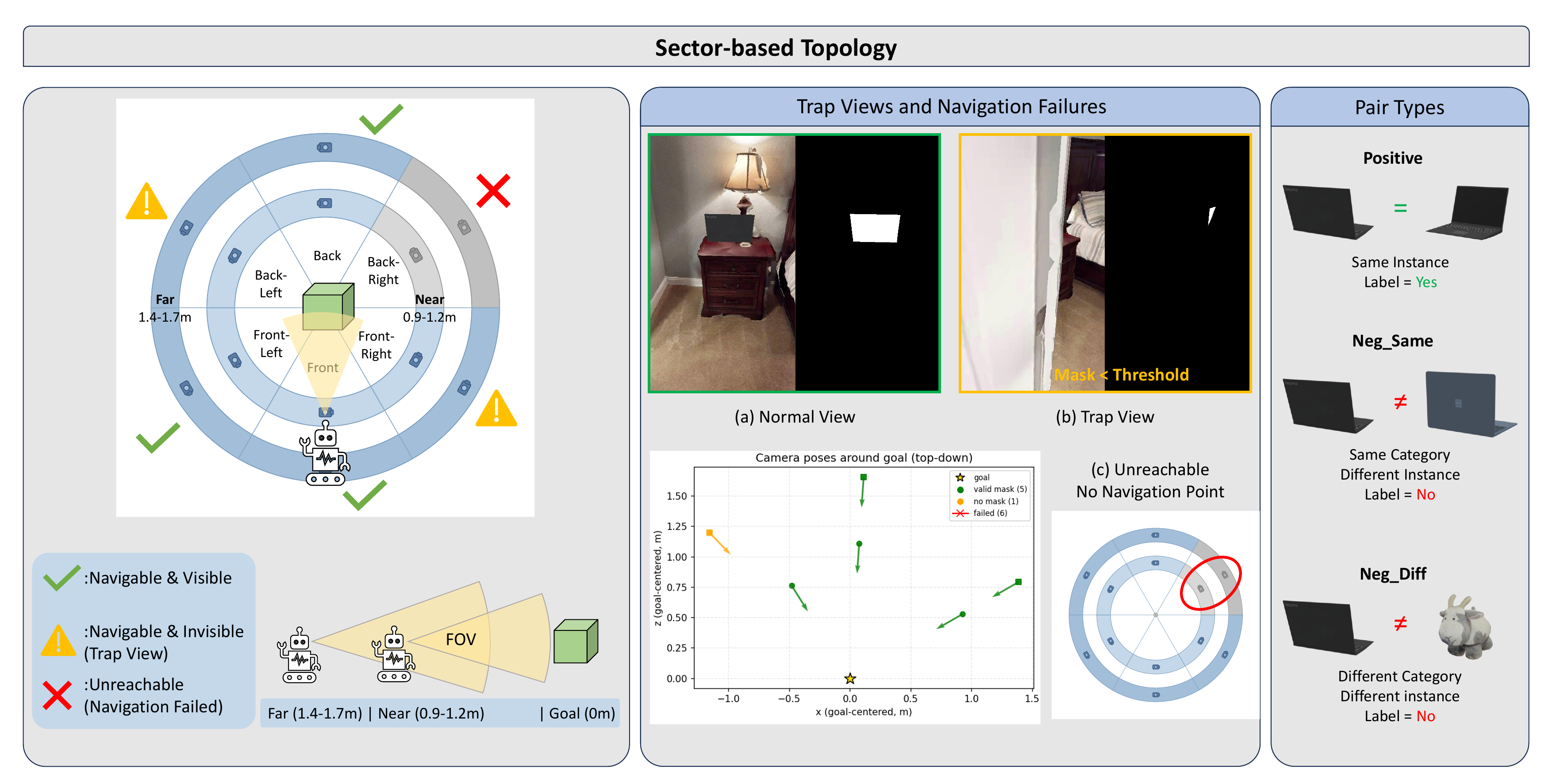}
  \caption{\textbf{Sector-based navigation topology and protocol-level annotations.} \textbf{Left:} 6 active sectors spaced every $60^\circ$ at two observation distances (far: 1.4--1.7\,m; near: 0.9--1.2\,m). Each sector is navigable~\&~visible (\textcolor{green!70!black}{\checkmark}), a \emph{trap view} (\textcolor{orange!80!black}{$\triangle$}, navigable but mask below threshold), or \emph{unreachable} (\textcolor{red}{$\times$}, no navigable point). \textbf{Center:} Examples of a normal view, a trap view, and an unreachable sector with corresponding top-down camera poses. \textbf{Right:} Three pair types used in evaluation: positive (same instance), neg\_same (same category, different instance), neg\_diff (different category).}
  \label{fig:sector_topology}
\end{figure*}

For each target object we construct an offline navigation graph by capturing multi-view observations under a controlled topology. Given the goal centroid $\mathbf{g} = (g_x, g_y, g_z)$, the azimuth of a capture position $\mathbf{c}$ in the horizontal plane is
\begin{equation}
  \theta = \mathrm{atan2}(c_z - g_z,\; c_x - g_x).
  \label{eq:atan2}
\end{equation}
We use $K = 6$ active azimuthal sectors, spaced every $60^\circ$ and defined relative to the starting capture's azimuth so the assignment is invariant to the absolute orientation choice. Each active sector is a $30^\circ$ angular window in the underlying 12-sector discretization (App.~\ref{app:capture_pipeline}). Captures are taken at two observation distances (far: 1.4--1.7\,m; near: 0.9--1.2\,m), with the camera oriented toward the goal. Each capture stores the RGB image, camera pose, instance segmentation mask, ground-truth bounding box, and a Boolean \texttt{mask\_meets\_threshold} indicating whether the projected mask exceeds a per-category visibility threshold. Sampling parameters are detailed in App.~\ref{app:capture_pipeline}; concrete capture episodes (full-visibility and trap-view) are shown in App.~\ref{app:capture_examples}. Figure~\ref{fig:sector_topology} summarizes the topology together with the trap/unreachable annotations (\S\ref{sec:trap}) and pair types (\S\ref{sec:pairs}).

\subsection{Trap views and unreachable sectors}
\label{sec:trap}

A distinguishing feature of \bench{} is the explicit modeling of two failure types that arise naturally in embodied perception but are not usually surfaced in purely geometric benchmarks. A capture $v$ is a \emph{trap view} if it is geometrically navigable but the target's projected mask falls below the visibility threshold $\tau$ (per-category values in App.~\ref{app:capture_pipeline}); trap views arise from self-occlusion, environmental occlusion, or extreme viewing angles. Other sectors are \emph{unreachable}: no navigable viewpoint exists (e.g., walls or furniture block access), and a \texttt{NAV}$_{d_k}$ to such a sector fails with the agent staying in place. Both failure types consume a step and are surfaced to the agent on the next observation via a \texttt{visibility\_warning} field. Sector indices in the capture data are sequential labels rather than fixed angular directions, so navigation actions are resolved geometrically: given a relative direction $d_k$ with target offset $\Delta\theta_k$, the environment computes the absolute target azimuth from the current pose and selects the closest capture within $\pm 30^\circ$ (otherwise the sector is declared unreachable). These annotations give \bench{} additional embodied-state structure beyond a static photo collection.

\subsection{Pair construction and dataset statistics}
\label{sec:pairs}

For each target object we sample (i)~a positive pair, (ii)~a neg\_same pair using a same-category distractor from the PInNED object pool, and (iii)~a neg\_diff pair using a cross-category distractor. From the 1{,}847 episodes that admit all three pair types we sample a balanced subset of 3{,}000 evaluation pairs (1{,}000 per type) spanning 71 unique target instances and 35 HM3D scenes. The full training capture comprises 37{,}583 raw episodes across 145 scenes and 264 instances; for our experiments we draw a disjoint 15{,}225-pair training pool from this set (App.~\ref{app:training_pool}). Table~\ref{tab:bench_overview} summarizes statistics.

\begin{table}[t]
\centering
\caption{\bench{} dataset statistics.}
\label{tab:bench_overview}
\small
\setlength{\tabcolsep}{4pt}
\begin{tabular}{l r r}
\toprule
\textbf{Property} & \textbf{Train} & \textbf{Eval} \\
\midrule
Raw captured episodes & 37{,}583 & 2{,}485 \\
Sampled paired episodes (1{:}1{:}1)\textsuperscript{$\dagger$} & 15{,}225 & 3{,}000 \\
Object categories & 18 & 18 \\
Unique object instances & 264 & 71 \\
HM3D scenes & 145 & 35 \\
Sectors per episode (max) & 6 & 6 \\
Observation rings & far, near & far, near \\
\bottomrule
\end{tabular}\\[2pt]
\raggedright\footnotesize $\dagger$ Subsample used in our experiments (5{,}075 episodes per pair type for training, 1{,}000 for evaluation). The full raw set is available for users who want to construct different training pools.
\end{table}

\subsection{Evaluation protocol}
\label{sec:metrics}

We report metrics along two axes.
\textbf{Accuracy}: overall, per-pair-type (positive / neg\_same / neg\_diff), and per-category.
\textbf{Efficiency \& diagnostics}: Average Steps to Decision (ASD) and Navigation Failure Rate. Together with the per-pair-type breakdown, these expose three failure axes that raw accuracy hides: calibration bias (confirmation vs.\ rejection, from the pair-type split), observation efficiency (ASD), and inability to recover from uninformative observations (Navigation Failure Rate). The evaluation harness additionally logs first-view accuracy and prediction-flip statistics for users who want finer-grained diagnostic views.

\subsection{Comparison with related benchmarks}
\label{sec:bench_compare}

Table~\ref{tab:bench_compare} positions \bench{} relative to navigation, grounding, and embodied-QA benchmarks. In this comparison, \bench{} is designed to cover all five axes we care about: (1)~active viewpoint selection, (2)~instance-level granularity, (3)~language-based queries, (4)~explicit multi-view reasoning, and (5)~offline reproducibility.

\begin{table}[t]
\centering
\caption{Comparison with related benchmarks. ``Act.'' = agent selects viewpoints; ``Inst.'' = instance-level targets; ``Lang.'' = language query; ``MV'' = multi-view reasoning; ``Off.'' = pre-captured.}
\label{tab:bench_compare}
\setlength{\tabcolsep}{2pt}
\small
\begin{tabular}{@{}l c c c c c l@{}}
\toprule
\textbf{Benchmark} & \textbf{Act.} & \textbf{Inst.} & \textbf{Lang.} & \textbf{MV} & \textbf{Off.} & \textbf{Task} \\
\midrule
ObjectNav~\cite{chaplot2020object} & \checkmark & $\times$ & $\times$ & $\times$ & $\times$ & Nav \\
OVON~\cite{yokoyama2024hm3d} & \checkmark & $\times$ & \checkmark & $\times$ & $\times$ & Nav \\
GOAT-Bench~\cite{khanna2024goat} & \checkmark & \checkmark & \checkmark & $\times$ & $\times$ & Nav \\
PIN~\cite{barsellotti2024personalized} & \checkmark & \checkmark & \checkmark & $\times$ & $\times$ & Nav \\
REVERIE~\cite{qi2020reverie} & \checkmark & \checkmark & \checkmark & $\times$ & $\times$ & Nav \\
RefCOCO~\cite{yu2016modeling} & $\times$ & \checkmark & \checkmark & $\times$ & \checkmark & Ground. \\
OpenEQA~\cite{majumdar2024openeqa}\textsuperscript{$\dagger$} & \checkmark & $\times$ & \checkmark & \checkmark & \checkmark & EQA \\
StaticEmbBench~\cite{xiao2025static} & $\times$ & $\times$ & \checkmark & $\times$ & \checkmark & VQA \\
VSI-Bench~\cite{yang2024thinking} & $\times$ & $\times$ & \checkmark & \checkmark & \checkmark & SpatialQA \\
EmbSpat.-B~\cite{du2024embspatial} & $\times$ & $\times$ & \checkmark & $\times$ & \checkmark & SpatialQA \\
3DSRBench~\cite{ma20253dsrbench} & $\times$ & $\times$ & \checkmark & $\times$ & \checkmark & SpatialQA \\
BLINK~\cite{fu2024blink} & $\times$ & $\times$ & \checkmark & \checkmark & \checkmark & PerceptQA \\
\midrule
\textbf{\bench{} (ours)} & \checkmark & \checkmark & \checkmark & \checkmark & \checkmark & \textbf{Verify} \\
\bottomrule
\end{tabular}\\[2pt]
\raggedright\footnotesize $\dagger$ OpenEQA's Active-EQA subtask supports active exploration; the Episodic-Memory-EQA mode is passive.
\end{table}

\paragraph{Why offline?}
Pre-captured observations avoid rendering stochasticity in evaluation and help isolate the verification decision from locomotion, which lets us run controlled ablations over detection, query mode, and NBV strategy~\cite{xiao2025static}. End-to-end integration with online navigation is left for future work.

\section{Reference Agents for \bench{}}
\label{sec:agent}

\begin{figure*}[t]
  \centering
  \includegraphics[width=\linewidth,trim=0 20 0 15,clip]{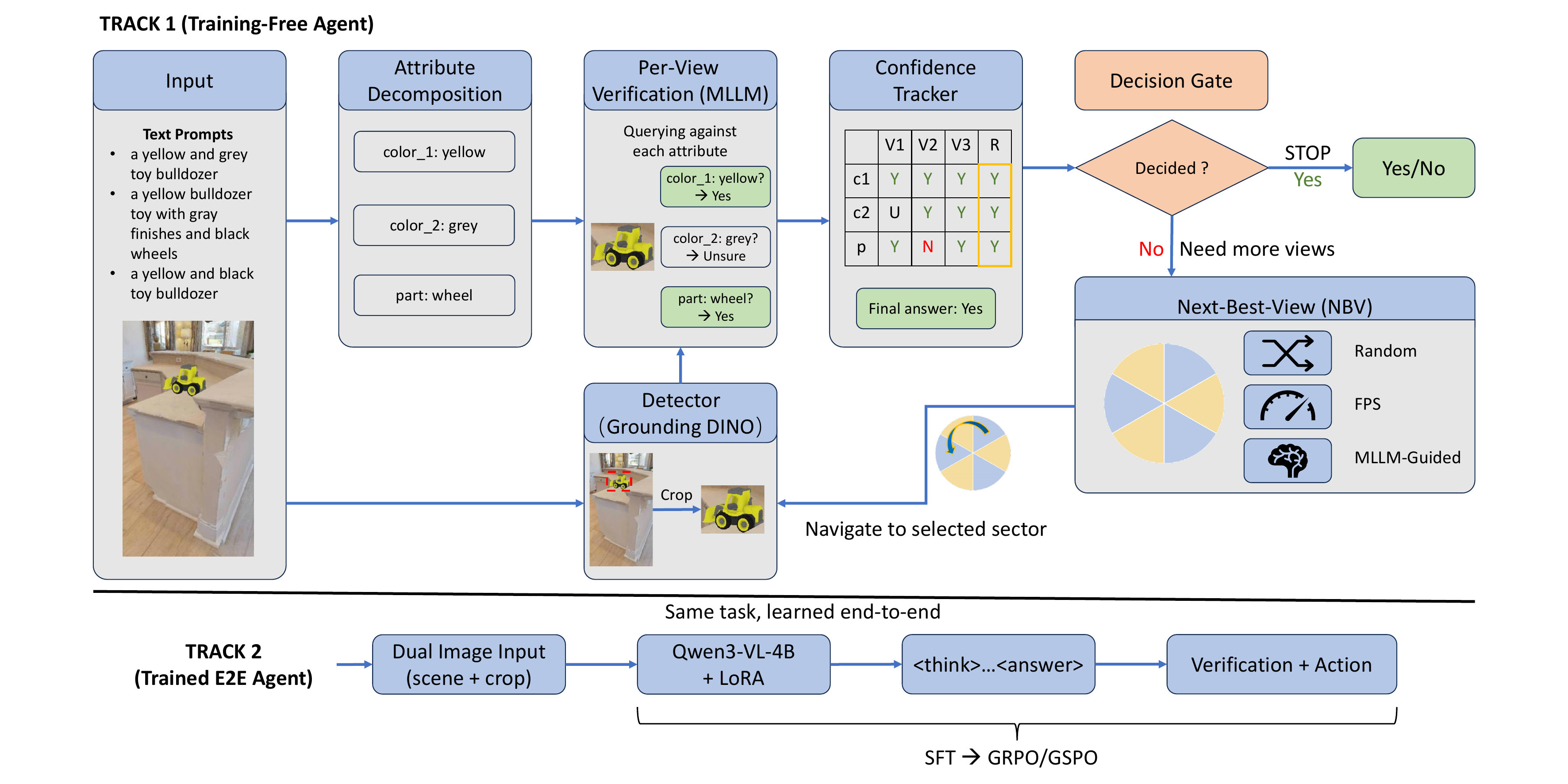}
  \caption{\textbf{Two reference agents.} Top: training-free modular pipeline with attribute decomposition, per-view verification, multi-view tracking, fusion, and next-best-view selection. Bottom: end-to-end LoRA-fine-tuned agent that performs verification and navigation in a single forward pass.}
  \label{fig:pipeline}
\end{figure*}

We build two reference agents whose results appear in \S\ref{sec:experiments}: a training-free modular pipeline and a LoRA-fine-tuned end-to-end agent. Both are intended as baselines for future work on \bench{}, not as method contributions, and are built from standard MLLM-prompting and fine-tuning components. In our ablations, confidence-weighted multi-view tracking with visibility-aware down-weighting helped reduce spurious rejections from low-visibility observations. We also adopt angular farthest-point sampling as a geometrically principled (not necessarily superior) NBV baseline. Figure~\ref{fig:pipeline} shows the pipeline; step-by-step walkthroughs of the verification process on four representative episodes (positive belief flip, navigation failure, correct distractor rejection, and trained-agent confirmation bias) are provided in App.~\ref{app:case_studies}.

\subsection{Training-free modular pipeline}

\paragraph{Attribute decomposition}
A text-only MLLM call first predicts the object category from $\mathcal{Q}$. The MLLM then decomposes $\mathcal{Q}$ into a structured list of up to $N_\text{max}=8$ attributes, each with a name, type, discriminative weight, and an evidence phrase quoted from $\mathcal{Q}$ (the quoted-evidence constraint suppresses hallucinated attributes). We compare against two ablations: \textbf{Direct} (the three descriptions are used holistically without decomposition) and \textbf{Merged} (the three descriptions are merged into a single sentence before holistic matching).

\paragraph{Per-view verification}
At each step the target object is localized (Grounding DINO~\cite{liu2024grounding} or, for analysis, ground-truth bounding boxes from the capture metadata) and cropped with a 3-pixel padding, upscaled if its shorter side is below 512\,px. For each as-yet-unresolved attribute, the MLLM is asked to return a state $s \in \{\texttt{Yes}, \texttt{No}, \texttt{Unsure}\}$ together with a free-text justification. Trap views are handled explicitly: when no GT mask is available, we feed the full scene image with detection confidence $c{=}0.1$, which propagates as low evidence weight in the tracker.

\paragraph{Confidence-weighted multi-view tracking}
For each attribute $a_i$ the tracker keeps a history $H_i = [(v_j, s_j, c_j)]$ of per-view outcomes. Setting $\tilde{c} = c$ if $c \ge \tau_\text{conf}{=}0.3$ and $\tilde{c} = 0.2\,c$ otherwise, we define for each state $\sigma$
\begin{equation}
  w_\sigma(a_i) = \!\!\!\sum_{(v,s,c)\in H_i,\, s=\sigma}\!\!\!\!\tilde{c},
  \quad \sigma\in\{\texttt{M}, \texttt{C}, \texttt{Mi}\},
  \label{eq:weights}
\end{equation}
where M, C, Mi abbreviate Matched, Contradictory, Missing. The reconciled state is Contradictory only if $w_\text{C}$ exceeds both $w_\text{M}$ and $w_\text{Mi}$; Matched if $w_\text{M} > 0.3$; Missing otherwise. The asymmetry is intentional: low-visibility observations (trap views push $\tilde{c}$ to $0.02$) cannot single-handedly trigger rejection, encoding the principle that unobserved is not contradicted. This conservative rule is intended to reduce spurious rejections on partially visible objects (\S\ref{sec:experiments}).

\paragraph{Fusion and adaptive stopping}
The reported multi-view runs use an adaptive attribute-majority fusion. After tracker reconciliation, \texttt{Missing} attributes abstain; the episode-level vote is \texttt{Yes} when matched attributes outnumber contradictions, \texttt{No} when contradictions outnumber matches, and conservative \texttt{No} on ties at the final step. Earlier in the episode, non-converged cases remain \texttt{Unsure}; the agent stops early only when the tracker has mathematically converged or all seen attributes are unanimous.

\paragraph{Next-best-view strategies}
When the fusion returns \texttt{Unsure}, the agent picks a next viewpoint. We compare three strategies. \textbf{Random} picks a uniformly random unvisited sector. \textbf{LLM-based NBV} prompts the MLLM with the current attribute states and a direction guide, asking it to select the most informative unvisited direction. \textbf{Angular FPS} picks the unvisited navigable direction $d_k$ whose azimuth $\theta_k$ maximizes the minimum shortest-arc distance to visited azimuths $\theta\in\Theta_\text{vis}$,
\begin{equation}
  d^\star = \argmax_{d_k}\,\min_{\theta\in\Theta_\text{vis}}\,\Delta(\theta_k,\theta),
  \label{eq:fps}
\end{equation}
where $\Delta(\alpha,\beta)=\min(|\alpha-\beta|,\,2\pi-|\alpha-\beta|)$ is the shortest-arc distance on the unit circle. Unlike Euclidean FPS on point clouds, Eq.~\eqref{eq:fps} is the correct metric for cameras arranged azimuthally on a horizontal ring around the object.

\subsection{Trained end-to-end agent}

We additionally fine-tune Qwen3-VL-4B~\cite{qwen2025qwen3vl} via LoRA~\cite{hu2022lora} (rank 16, $\alpha{=}32$) as an end-to-end agent that consumes the scene RGB plus a candidate-object crop and emits a structured \texttt{<think>}/\texttt{<answer>} response with the verification decision and the next action in a single forward pass. Training always uses GT bounding boxes to produce the crop; at inference, the crop is produced either by Grounding DINO (\textbf{DINO} mode) or by ground-truth bounding boxes from the capture metadata (\textbf{GT} mode), matching the two evaluation settings reported in \S\ref{sec:experiments}. We train on $\sim$22K samples whose chain-of-thought labels are templated directly from ground-truth pair information and per-step visibility metadata supplied by the benchmark (\textbf{SFT}; schema in App.~\ref{app:sft_format}). On top of the SFT adapter we explore three post-SFT alignment strategies: offline preference optimization \textbf{DPO}~\cite{rafailov2024direct}, online \textbf{GRPO}~\cite{shao2024deepseekmath} with $G{=}4$ completions per prompt, and its sequence-level variant \textbf{GSPO}~\cite{zheng2025gspo}. The composite reward is verification correctness (soft partial credit) plus action quality (a 4-tier scheme that consumes the benchmark's per-episode \texttt{visible}/\texttt{navigable}/\texttt{best\_sectors} metadata, App.~\ref{app:rl_format}) plus format compliance. Both labels and reward thus depend on benchmark-side ground truth, intentional for a reference baseline. Hyperparameters and the full reward specification are in App.~\ref{app:training_details}.

\section{Experiments}
\label{sec:experiments}

\subsection{Setup}
\label{sec:setup}

All evaluations use the 3{,}000-episode \bench{} test set with adaptive stopping over $T{=}6$ steps. The MLLM backbone is Qwen3-VL-4B~\cite{qwen2025qwen3vl} unless otherwise stated. We also evaluate Qwen3-VL-8B and SenseNova-SI-1.2-InternVL3-8B~\cite{zhu2025internvl3, wang2025sensenova} for cross-model comparison, and CLIP~\cite{radford2021learning} and SigLIP2~\cite{tschannen2025siglip} as embedding-only baselines. The default detection frontend is Grounding DINO~\cite{liu2024grounding}; GT bounding boxes are used in controlled ablations. Trained agents are LoRA fine-tuned (rank 16, $\alpha{=}32$) on $\sim$22K SFT samples from the disjoint training pool, with optional DPO, GRPO, or GSPO post-SFT alignment (\S\ref{sec:agent}); hyperparameters are in App.~\ref{app:training_details}. The 95\% binomial confidence interval at $p{=}0.85$ on $n{=}3{,}000$ episodes is $\approx \pm 1.3$\,pp; we treat smaller differences as functionally equivalent. We bold the best Pos accuracy alongside the best Overall because Pos is the harder axis; when a table mixes DINO and GT modes, bolding is applied within each detection setting.

\subsection{Training-free results}
\label{sec:tf_results}

Table~\ref{tab:main_results} reports training-free results on Qwen3-VL-4B with Grounding DINO detection.

\begin{table}[t]
\centering
\caption{Training-free results on \bench{} (Qwen3-VL-4B, DINO detection, 3{,}000 episodes, adaptive stopping). 95\% binomial CI at $p{=}0.85$ is $\pm$1.3\,pp.}
\label{tab:main_results}
\small
\setlength{\tabcolsep}{2pt}
\begin{tabular}{@{}l c c c c c@{}}
\toprule
\textbf{Method} & \textbf{Overall} & \textbf{Pos} & \textbf{Neg\_Same} & \textbf{Neg\_Diff} & \textbf{ASD} \\
\midrule
\multicolumn{6}{l}{\textit{Single-View}} \\
\quad Attr & 0.844 & \textbf{0.652} & 0.918 & 0.961 & 1.0 \\
\quad Direct & 0.813 & 0.457 & 0.984 & 0.999 & 1.0 \\
\quad Merged & 0.815 & 0.468 & 0.982 & 0.996 & 1.0 \\
\midrule
\multicolumn{6}{l}{\textit{Multi-View, attribute decomposition}} \\
\quad + Random & 0.849 & 0.592 & 0.965 & 0.989 & 2.34 \\
\quad + Angular FPS & 0.848 & 0.592 & 0.964 & 0.989 & 2.36 \\
\quad + LLM-NBV & \textbf{0.850} & 0.596 & 0.965 & 0.988 & 2.34 \\
\midrule
\multicolumn{6}{l}{\textit{Multi-View, direct query}} \\
\quad + Random & 0.827 & 0.492 & 0.991 & 0.999 & 2.09 \\
\quad + Angular FPS & 0.826 & 0.490 & 0.990 & 0.999 & 2.13 \\
\quad + LLM-NBV & 0.825 & 0.486 & 0.990 & 0.999 & 2.08 \\
\bottomrule
\end{tabular}
\end{table}

\paragraph{Attribute decomposition helps in these settings}
Across single-view and multi-view, attribute decomposition is 2.5--3.1\,pp higher than direct query overall, with the gap concentrated on positive pairs (e.g., 0.652 vs.\ 0.457 in single-view). This is consistent with decomposition encouraging per-attribute confirmation and reducing the rejection bias of holistic matching.

\paragraph{Multi-view helps mostly for negatives}
Single-view Attr already reaches 0.844; the best multi-view configuration reaches 0.850, a 0.6\,pp gain within the 95\% CI. The improvement is uneven: neg\_same goes from 0.918 to 0.965 (+4.7\,pp) while positive accuracy drops from 0.652 to 0.596. The drop also appears with GT bounding boxes (0.758 to 0.728), so it is not explained solely by DINO detection. Under the conservative confidence-weighted tracker, more views give more chances for an attribute to register as not confirmed, tightening the bar for positive verification while strengthening rejection.

\paragraph{NBV strategies are within CI resolution at this scale}
Random, Angular FPS, and LLM-NBV land within 0.2\,pp of each other on Qwen3-VL-4B (0.848--0.850), well inside the 95\% CI; the same within-CI pattern holds for the other tested on-device MLLMs (App.~\ref{app:full_results}). Under this prompting setup, any gain from smarter view selection appears to be swamped by single-step verification noise. The three strategies also incur similar navigation-failure counts (Random 1{,}793; FPS 1{,}958; LLM 1{,}787; full breakdown in App.~\ref{app:nbv_diag}), suggesting that per-step evidence extraction is a major limiting factor in these baselines. \bench{} leaves room for future agents with stronger uncertainty awareness to compete on this dimension.

\subsection{Cross-model and embedding baselines}
\label{sec:cross_model}

Table~\ref{tab:cross_model} compares MLLM backbones with two embedding-only baselines.

\begin{table}[t]
\centering
\caption{Cross-model comparison (DINO detection, 3{,}000 episodes). Best configuration per model by Overall, with ASD used to break ties. SenseNova-SI here denotes SenseNova-SI-1.2-InternVL3-8B.}
\label{tab:cross_model}
\small
\setlength{\tabcolsep}{2pt}
\begin{tabular}{@{}l l c c c c c@{}}
\toprule
\textbf{Model} & \textbf{Cfg} & \textbf{Ovr.} & \textbf{Pos} & \textbf{N\_S} & \textbf{N\_D} & \textbf{ASD} \\
\midrule
\multicolumn{7}{l}{\textit{Embedding-based}} \\
CLIP & SV-Merged & 0.771 & 0.349 & 0.965 & 1.000 & 1.0 \\
SigLIP2 & SV-Merged & 0.801 & 0.429 & 0.974 & 1.000 & 1.0 \\
\midrule
\multicolumn{7}{l}{\textit{MLLM-based}} \\
Qwen3-VL-4B & MV-Attr+LLM & \textbf{0.850} & 0.596 & 0.965 & 0.988 & 2.34 \\
Qwen3-VL-8B & MV-Attr+LLM & 0.797 & 0.412 & 0.979 & 1.000 & 1.94 \\
SenseNova-SI & MV-Direct+Rnd & 0.833 & \textbf{0.658} & 0.884 & 0.958 & 1.62 \\
\bottomrule
\end{tabular}
\end{table}

\paragraph{Best on-device MLLM exceeds best embedding; calibration affects query mode}
Among the tested on-device MLLM baselines, the best one (0.850) exceeds the best embedding baseline (SigLIP2, 0.801) by +4.9\,pp, with a much larger gap on positive pairs (0.596 vs.\ 0.429). CLIP and SigLIP2 sustain near-perfect neg\_same accuracy with a strong ``no match'' bias, but in our embedding-only implementation they do not perform explicit attribute reasoning, and their positive accuracy remains much lower than the best MLLM baseline. Among MLLMs, Qwen3-VL-8B is more rejection-biased than 4B (0.979 vs.\ 0.965 on neg\_same, 0.412 vs.\ 0.596 on positive), and attribute decomposition, which requires confirming every attribute, performs worse for it; SenseNova-SI shows the opposite pattern, with neg\_same dropping to 0.526 under attribute decomposition because it confirms too readily, while Direct query reaches 0.833 overall. These results suggest that the best-performing query strategy depends on the model's calibration profile, not just on its raw capability.

\subsection{Trained agents}
\label{sec:trained_results}

\begin{table}[t]
\centering
\caption{Trained agents on \bench{} (Qwen3-VL-4B + LoRA, 3{,}000 episodes). TF-Best is MV-Attr+LLM (DINO). NavFail = navigation failure rate (\%).}
\label{tab:trained_results}
\small
\setlength{\tabcolsep}{2pt}
\begin{tabular}{@{}l l c c c c c c@{}}
\toprule
\textbf{Method} & \textbf{Det.} & \textbf{Ovr.} & \textbf{Pos} & \textbf{N\_S} & \textbf{N\_D} & \textbf{ASD} & \textbf{NavF} \\
\midrule
TF-Best & DINO & 0.850 & 0.596 & 0.965 & 0.988 & 2.34 & 28.1 \\
\midrule
Base (no FT) & DINO & 0.706 & 0.146 & 0.973 & 0.999 & 1.74 & 16.3 \\
SFT & DINO & 0.848 & \textbf{0.759} & 0.814 & 0.971 & 1.96 & 18.2 \\
SFT+GRPO & DINO & 0.853 & 0.736 & 0.838 & 0.985 & 1.61 & 9.0 \\
SFT+GSPO & DINO & \textbf{0.856} & 0.745 & 0.839 & 0.985 & 1.62 & 9.1 \\
SFT+GSPO & GT & \textbf{0.889} & \textbf{0.813} & 0.864 & 0.991 & 1.65 & 9.9 \\
\bottomrule
\end{tabular}
\end{table}

Table~\ref{tab:trained_results} reports trained-agent results. The un-tuned Qwen3-VL-4B reaches only 0.706 overall (Pos 0.146), with a strong ``no match'' bias. SFT closes most of the gap (0.848, Pos 0.759, $+$61.3\,pp); GRPO and GSPO add $\sim$1\,pp overall and reduce NavFail from 18.2\% to 9.0--9.1\% in this setup. With GT, SFT+GSPO reaches 0.889. DPO under a Specific-CoT variant (App.~\ref{app:full_results}) is essentially indistinguishable from SFT. The two paradigms trade off calibration: TF-Best is higher on negative rejection (neg\_same 0.965 vs.\ 0.839), while trained agents are higher on positive confirmation (Pos $+$14.9\,pp) and efficiency (ASD 1.62 vs.\ 2.34, NavFail 9\% vs.\ 28\%); a scatter plot of all methods on the accuracy--efficiency plane is in App.~\ref{app:eff_acc}.

\subsection{Detection quality and per-category breakdown}
\label{sec:detection_bottleneck}
\label{sec:per_category}

For SV-Attr, GT-bbox mode reaches 0.875 vs.\ 0.844 with DINO (+3.1\,pp), gain concentrated on positive pairs (0.758 vs.\ 0.652, +10.6\,pp); the SFT+GSPO GT/DINO gap is +3.3\,pp. The GT-box ablations identify detection quality as a measured bottleneck on \bench{}. App.~\ref{app:pos_failure}/\ref{app:det_quality} decompose positive failures (rejection-bias 53--94\%, trap/obs 6--34\%) and report a Pos--detection-confidence correlation ($r{=}0.932$, small-$n$). Per-category numbers (heatmap and full table) are in App.~\ref{app:per_category}: difficulty appears correlated with category scale, with small categories (wallet, watch, keys) at roughly 0.68--0.72 overall under MV-Attr+LLM and large distinctive ones (backpack, teddy bear) above 0.92; TF-Best is higher on neg\_same rejection while SFT+GSPO recovers small-object positives.

\section{Discussion and Conclusion}
\label{sec:discussion}

\bench{} isolates semantic verification between navigation and interaction. On our on-device MLLM baselines ($\leq$8B), the best MLLM baseline exceeds the best embedding baseline by 4.9\,pp, GT-box ablations show a +3.1--3.3\,pp detection gap, and three NBV strategies fall within 95\% CI without reliable gains from active selection.

A LoRA-fine-tuned agent (SFT+GSPO) reaches 85.6\% (88.9\% with GT), trading neg\_same accuracy for stronger positive confirmation and lower NavFail (9\% vs.\ 28\%); the complementary failure modes suggest calibration matters at least as much as raw capability for this task.

{
    \small
    \bibliographystyle{ieeenat_fullname}
    \bibliography{main}
}

\appendix
%

\clearpage
\onecolumn

\section*{Supplementary Material}

App.~\ref{app:capture_pipeline} gives the multi-view capture pipeline; App.~\ref{app:capture_examples} a trap-view capture example; App.~\ref{app:prompts} the reference-agent prompt templates; App.~\ref{app:index_generation} the evaluation-index construction; App.~\ref{app:capture_stats} capture statistics; App.~\ref{app:training_pool} the training-pool construction; App.~\ref{app:training_details} the high-level training configuration; App.~\ref{app:data_format} the on-disk data formats; App.~\ref{app:eff_acc} the efficiency--accuracy scatter; App.~\ref{app:per_category} the per-category breakdown; App.~\ref{app:diagnostics} four diagnostic tables (NBV navigation failures, NBV efficiency, positive-failure root-cause decomposition, per-category detection quality); App.~\ref{app:case_studies} four qualitative case studies of the verification process; App.~\ref{app:full_results} the complete results.


\section{Capture Pipeline Specification}
\label{app:capture_pipeline}

We document the multi-view capture pipeline used to build \bench{}.


\subsection{Camera and Sensor Configuration}
\label{app:camera_config}

Table~\ref{tab:app_camera_params} lists the camera and agent parameters used throughout the capture process.

\begin{table}[htbp]
\centering
\caption{Camera and agent parameters for multi-view capture.}
\label{tab:app_camera_params}
\begin{tabular}{l l}
\toprule
\textbf{Parameter} & \textbf{Value} \\
\midrule
Image resolution (W $\times$ H) & 360 $\times$ 640 (portrait) \\
Horizontal field of view (HFOV) & 42.0\textdegree \\
Vertical field of view (VFOV) & 68.6\textdegree \\
Focal length ($f_x = f_y$) & $\approx$ 469 px \\
Principal point ($c_x, c_y$) & (179.5, 319.5) \\
RGB sensor height & 1.31 m \\
Agent body height & 1.41 m \\
Agent collision radius & 0.17 m \\
\bottomrule
\end{tabular}
\end{table}

The vertical field of view is derived from the horizontal FOV and the aspect ratio:
\begin{equation}
\label{eq:vfov}
\text{VFOV} = 2 \cdot \arctan\!\left(\tan\!\left(\frac{\text{HFOV}}{2}\right) \cdot \frac{H}{W}\right)
= 2 \cdot \arctan\!\left(\tan(21^\circ) \cdot \frac{640}{360}\right) \approx 68.6^\circ.
\end{equation}

The camera intrinsic matrix is:
\begin{equation}
\label{eq:intrinsics}
\mathbf{K} = \begin{bmatrix} f_x & 0 & c_x \\ 0 & f_y & c_y \\ 0 & 0 & 1 \end{bmatrix},
\quad \text{where}\quad
f_x = \frac{W}{2\,\tan(\text{HFOV}/2)}, \quad
f_y = \frac{H}{2\,\tan(\text{VFOV}/2)}.
\end{equation}
With the given parameters, $f_x = f_y \approx 469$ pixels, confirming square pixel geometry.


\subsection{Sector Division and Coordinate System}
\label{app:sector_division}

The azimuthal plane around each target object is divided into $N = 12$ sectors of $30^\circ$ each. With a sector skip parameter of 1, the capture pipeline uses 6 \emph{active} sectors at indices $\{0, 2, 4, 6, 8, 10\}$, effectively spaced $60^\circ$ apart.

\paragraph{Polar coordinate computation.}
Given a camera position $\mathbf{c} = (c_x, c_y, c_z)$ and a goal position $\mathbf{g} = (g_x, g_y, g_z)$, the azimuthal angle and radial distance in the horizontal XZ plane are:
\begin{align}
\label{eq:polar}
r &= \sqrt{(c_x - g_x)^2 + (c_z - g_z)^2}, \\
\theta &= \text{atan2}(c_z - g_z, \; c_x - g_x),
\end{align}
where $\theta$ is normalized to $[0, 2\pi)$ by adding $2\pi$ when negative. The angle is measured counter-clockwise from the positive $x$-axis in the XZ plane; the $y$-axis is vertical (up) in Habitat's coordinate system. All angle computations in this paper are performed in radians; degree values shown elsewhere (e.g., the $30^\circ$ sector width and Table~\ref{tab:active_sectors}) are equivalent labels intended for readability.

\paragraph{Sector membership.}
The 12 sector boundaries are:
\begin{equation}
S_i = [i \cdot \tfrac{\pi}{6}, \; (i+1) \cdot \tfrac{\pi}{6}), \quad i = 0, 1, \ldots, 11.
\end{equation}
A viewpoint at angle $\theta$ belongs to sector $S_i$ if $\theta \in S_i$.
Table~\ref{tab:active_sectors} lists the six active sectors and their angular ranges. The evaluation index files and downstream code use the raw active indices $\{0, 2, 4, 6, 8, 10\}$ directly; the main text's $K{=}6$ refers to the cardinality of the active set rather than to a contiguous 0--5 range.

\begin{table}[htbp]
\centering
\caption{Active sector indices and angular ranges (with sector\_skip\,=\,1).}
\label{tab:active_sectors}
\begin{tabular}{c c}
\toprule
\textbf{Sector index} & \textbf{Angular range} \\
\midrule
0  & $[0^\circ, 30^\circ)$ \\
2  & $[60^\circ, 90^\circ)$ \\
4  & $[120^\circ, 150^\circ)$ \\
6  & $[180^\circ, 210^\circ)$ \\
8  & $[240^\circ, 270^\circ)$ \\
10 & $[300^\circ, 330^\circ)$ \\
\bottomrule
\end{tabular}
\end{table}


\subsection{Distance Ranges}
\label{app:distance_ranges}

For each active sector, the pipeline attempts to capture one \emph{far} view and one \emph{near} view, yielding up to 12 viewpoints per episode ($6 \text{ sectors} \times 2 \text{ distance levels}$).

\begin{table}[htbp]
\centering
\caption{Distance range parameters.}
\label{tab:distance_ranges}
\begin{tabular}{l c c}
\toprule
\textbf{Range type} & \textbf{Min (m)} & \textbf{Max (m)} \\
\midrule
Near & 0.9 & 1.2 \\
Far  & 1.4 & 1.7 \\
\bottomrule
\end{tabular}
\end{table}

The 0.2\,m gap between near max (1.2\,m) and far min (1.4\,m) keeps the two ranges disjoint. Far views give scene context; near views give surface detail. Ranges are horizontal (XZ-plane) distance from camera to goal; the 3D Euclidean distance differs for elevated objects.


\subsection{Viewpoint Sampling Algorithm}
\label{app:sampling_algorithm}

Algorithm~\ref{alg:sampling} gives the sampling procedure. Far views are sampled before near views so each near view can be aligned with its corresponding far view, producing a natural zoom-in pair.

\begin{algorithm}[htbp]
\caption{Multi-view viewpoint sampling for one episode.}
\label{alg:sampling}
\begin{algorithmic}[1]
\Require Goal position $\mathbf{g}$; active sectors $\mathcal{S} = \{0, 2, 4, 6, 8, 10\}$; distance ranges $[r_{\min}^{\text{near}}, r_{\max}^{\text{near}}]$, $[r_{\min}^{\text{far}}, r_{\max}^{\text{far}}]$; max attempts $T = 50$; consecutive failure limit $F = 15$
\Ensure Set of viewpoint records $\mathcal{V}$
\State Initialize per-episode RNG seed via Eq.~\eqref{eq:seed} (\S\ref{app:seeding})
\For{each sector $s \in \mathcal{S}$}
    \State $[\alpha_{\min}, \alpha_{\max}] \gets$ angular bounds of sector $s$

    \Statex \Comment{\textbf{--- Far view ---}}
    \State $\theta_{\text{far}} \gets \texttt{null}$; $\texttt{consec\_fail} \gets 0$
    \For{$t = 1$ to $T$}
        \State Sample $r \sim \mathcal{U}(r_{\min}^{\text{far}}, r_{\max}^{\text{far}})$
        \State $\mathbf{p} \gets$ \Call{RingSample}{$\mathbf{g}, r, [\alpha_{\min}, \alpha_{\max}]$}
        \If{$\mathbf{p}$ is not navigable}
            \State $\texttt{consec\_fail} \gets \texttt{consec\_fail} + 1$
            \If{$\texttt{consec\_fail} \geq F$} \textbf{break} \EndIf
            \State \textbf{continue}
        \EndIf
        \State $\texttt{consec\_fail} \gets 0$
        \State Verify $\mathbf{p}$ is within sector $s$ and distance range
        \State Orient camera toward $\mathbf{g}$; apply pitch adjustment (\S\ref{app:pitch})
        \If{\Call{FrustumCheck}{camera, $\mathbf{g}$} = \texttt{true}} \Comment{\S\ref{app:frustum}}
            \State Capture RGB + semantic observation; record viewpoint
            \State $\theta_{\text{far}} \gets \text{atan2}(c_z - g_z, c_x - g_x)$
            \State \textbf{break}
        \EndIf
    \EndFor

    \Statex \Comment{\textbf{--- Near view (aligned with far view) ---}}
    \State $\texttt{ray\_attempts} \gets 0$
    \For{$t = 1$ to $T$}
        \State Sample $r \sim \mathcal{U}(r_{\min}^{\text{near}}, r_{\max}^{\text{near}})$
        \State $\mathbf{p} \gets \texttt{null}$
        \If{$\theta_{\text{far}} \neq \texttt{null}$ \textbf{and} $\texttt{ray\_attempts} < 3$}
            \State $\mathbf{p} \gets$ \Call{RaySample}{$\mathbf{g}, r, \theta_{\text{far}}, \pm 2^\circ$}; $\texttt{ray\_attempts} \mathrel{+}= 1$ \Comment{Align with far view}
        \EndIf
        \If{$\mathbf{p} = \texttt{null}$}
            \State $\mathbf{p} \gets$ \Call{RingSample}{$\mathbf{g}, r, [\alpha_{\min}, \alpha_{\max}]$}  \Comment{Fallback}
        \EndIf
        \State Apply same navigability, sector, frustum checks as far view
        \If{capture succeeds} \textbf{break} \EndIf
    \EndFor
\EndFor
\State \Return $\mathcal{V}$
\end{algorithmic}
\end{algorithm}

\paragraph{Ring sampling.}
Given goal position $\mathbf{g}$, radius $r$, and angular bounds $[\alpha_{\min}, \alpha_{\max}]$:
\begin{enumerate}[leftmargin=2em]
    \item Sample $\alpha \sim \mathcal{U}(\alpha_{\min}, \alpha_{\max})$.
    \item Compute candidate: $x = g_x + r \cos \alpha$, $z = g_z + r \sin \alpha$.
    \item Snap to the nearest navigable point on the NavMesh via the Habitat pathfinder.
    \item Accept if the snapped point is navigable; reject otherwise.
\end{enumerate}

\paragraph{Ray-aligned sampling.}
To align the near view with the far view's direction $\theta_{\text{far}}$:
\begin{enumerate}[leftmargin=2em]
    \item On the first attempt, use $\theta = \theta_{\text{far}}$ exactly.
    \item On subsequent attempts, add a small perturbation: $\theta = \theta_{\text{far}} + \delta$, where $\delta \sim \mathcal{U}(-2^\circ, +2^\circ)$.
    \item Compute candidate: $x = g_x + r \cos \theta$, $z = g_z + r \sin \theta$.
    \item Snap and validate as in ring sampling.
\end{enumerate}
Up to 3 ray-aligned attempts are made before falling back to unconstrained ring sampling within the sector.


\subsection{Frustum Visibility Check}
\label{app:frustum}

After positioning the agent and orienting the camera toward the goal, a frustum check verifies that the goal position falls within the camera's field of view. Given the camera's orthonormal axes $(\mathbf{f}, \mathbf{r}, \mathbf{u})$ (forward, right, up) and position $\mathbf{p}$, the offset vector $\mathbf{v} = \mathbf{g} - \mathbf{p}$ is projected along each axis:
\begin{align}
d &= \mathbf{f} \cdot \mathbf{v} \quad \text{(depth)}, \\
r_x &= \mathbf{r} \cdot \mathbf{v} \quad \text{(horizontal offset)}, \\
r_y &= \mathbf{u} \cdot \mathbf{v} \quad \text{(vertical offset)}.
\end{align}
The goal is considered visible if and only if:
\begin{equation}
\label{eq:frustum}
d_{\min} < d < d_{\max} \quad \wedge \quad |r_x| \leq d \cdot \tan\!\left(\frac{\text{HFOV}}{2}\right) \quad \wedge \quad |r_y| \leq d \cdot \tan\!\left(\frac{\text{VFOV}}{2}\right),
\end{equation}
where $d_{\min} = 0.05$\,m (near clip) and $d_{\max} = 100$\,m (far clip). With the configured FOV values, the half-angle tangents are $\tan(21^\circ) \approx 0.384$ (horizontal) and $\tan(34.3^\circ) \approx 0.683$ (vertical).


\subsection{Camera Pitch Adjustment}
\label{app:pitch}

After yaw alignment (orienting the camera to face the goal in the horizontal plane), the pipeline checks whether a vertical pitch adjustment is needed. Let $h_{\text{diff}} = g_y - (b_y + h_{\text{sensor}})$, where $b_y$ is the agent's base height (from the NavMesh floor) and $h_{\text{sensor}} = 1.31$\,m is the sensor offset. The pitch adjustment rule is:
\begin{equation}
\text{action} = \begin{cases}
\texttt{look\_up}   & \text{if } h_{\text{diff}} > \tau, \\
\texttt{look\_down} & \text{if } h_{\text{diff}} < -\tau, \\
\text{none}         & \text{otherwise},
\end{cases}
\end{equation}
where $\tau = 0.3$\,m is the pitch threshold. Each pitch action rotates the camera by one discrete step (as defined by the Habitat agent configuration). The action taken is recorded in the viewpoint metadata.


\subsection{Per-Category Mask Thresholds}
\label{app:mask_thresholds}

A viewpoint's \emph{visibility quality} is assessed by counting the number of pixels in the rendered semantic segmentation map that correspond to the target object (semantic ID = 100000). A viewpoint is considered to have a valid mask if this pixel count exceeds a category-specific threshold. Table~\ref{tab:mask_thresholds} lists the thresholds organized by tier.

\begin{table}[htbp]
\centering
\caption[Per-category mask area thresholds]{Per-category mask area thresholds (in pixels). All thresholds are calibrated for the 360$\times$640 capture resolution at the near/far distance ranges.}
\label{tab:mask_thresholds}
\begin{tabular}{c c l}
\toprule
\textbf{Tier} & \textbf{Threshold (px)} & \textbf{Categories} \\
\midrule
S & 100 & keys, watch \\
A & 150 & eyeglasses, wallet, cellphone, visor, camera, mug \\
B & 300 & toy, ball, headphones, hat, book, shoes \\
C & 500 & backpack, bag, laptop, teddy bear \\
\midrule
Default & 200 & (fallback for unlisted categories) \\
\bottomrule
\end{tabular}
\end{table}

Thresholds are tiered by object size: small objects (keys, watch) project fewer pixels even at near distance, so a lower threshold avoids systematically dropping them; large objects (backpack, laptop) cover more area, so a higher threshold filters out marginal views.


\subsection{Episode Success Criteria}
\label{app:success_criteria}

An episode is considered \emph{successful} (included in the final dataset) if both of the following conditions are met:
\begin{align}
\text{navigable\_count} &\geq 6 \quad \text{(out of 12 total viewpoints)}, \label{eq:nav_thresh} \\
\text{valid\_mask\_count} &\geq 3 \quad \text{(viewpoints with mask area $\geq$ category threshold)}. \label{eq:mask_thresh}
\end{align}
\texttt{navigable\_count} counts viewpoints (max 12 = 6 sectors $\times$ 2 ranges), not sectors (max 6). \texttt{valid\_mask\_count} counts viewpoints whose mask exceeds the category threshold. The evaluation pipeline reports ``navigable sectors'' (Fig.~\ref{fig:app_sector_dist}), counting sectors with at least one navigable viewpoint; this is a sector-level number and differs from the viewpoint-level count here.


\subsection{Deterministic Per-Episode Seeding}
\label{app:seeding}

To ensure reproducibility across distributed jobs and pipeline iterations, each episode uses a deterministic random seed derived from its identity:
\begin{equation}
\label{eq:seed}
\texttt{seed}_{\text{ep}} = \text{MD5}\!\left(\texttt{scene\_id} \| \text{``\_''} \| \texttt{episode\_id} \| \text{``\_''} \| \texttt{base\_seed}\right) \bmod 2^{31},
\end{equation}
where $\texttt{base\_seed} = 42$ by default. Both NumPy and Python's \texttt{random} module are seeded with $\texttt{seed}_{\text{ep}}$ before sampling viewpoints for that episode.

The same episode produces identical viewpoints across workers and across pipeline re-runs, while different episodes within a scene receive different sequences. The base seed (not a job-offset seed) must be used to keep cross-worker outputs identical.

\subsection{Dataset-Level Statistics}
\label{app:rendering}

After applying the success criteria of \S\ref{app:success_criteria} across all captured scenes, the final validation set contains 2{,}485 episodes, of which 1{,}847 are eligible for evaluation index generation (App.~\ref{app:index_generation}). Three object instances were excluded during quality assurance due to renderer-level texture anomalies in the underlying simulator; the remaining 335 instances cover all 18 categories.

\section{Multi-View Capture: Concrete Episode Examples}
\label{app:capture_examples}

\begin{figure}[ht]
  \centering
  \includegraphics[width=\linewidth]{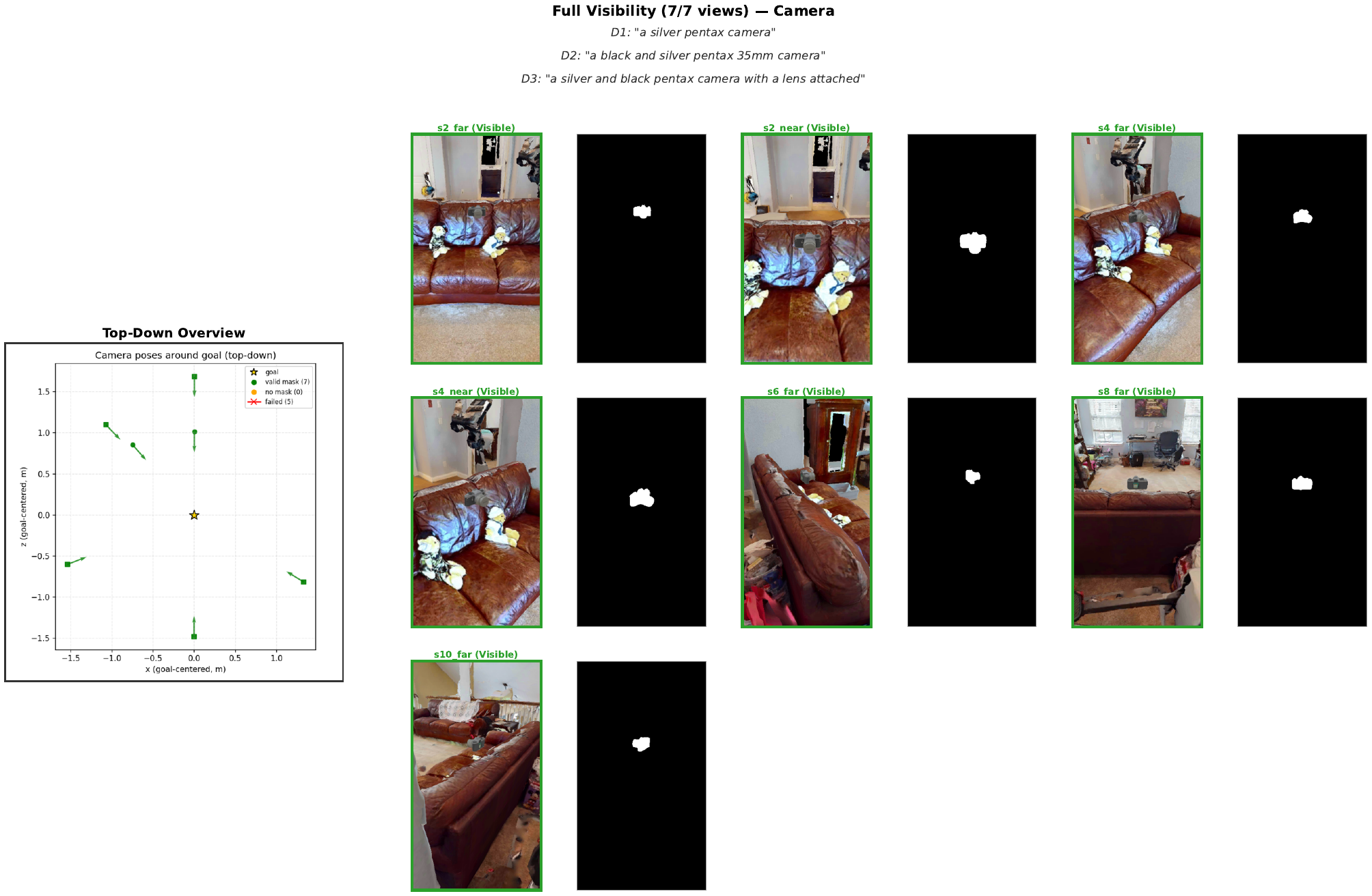}
  \caption{\textbf{Multi-view capture (full visibility).} Top-down camera-pose view (left) and per-sector RGB observations with paired masks (right). All navigable sectors yield valid masks above the per-category threshold.}
  \label{fig:capture_full}
\end{figure}

\begin{figure}[ht]
  \centering
  \includegraphics[width=0.85\linewidth]{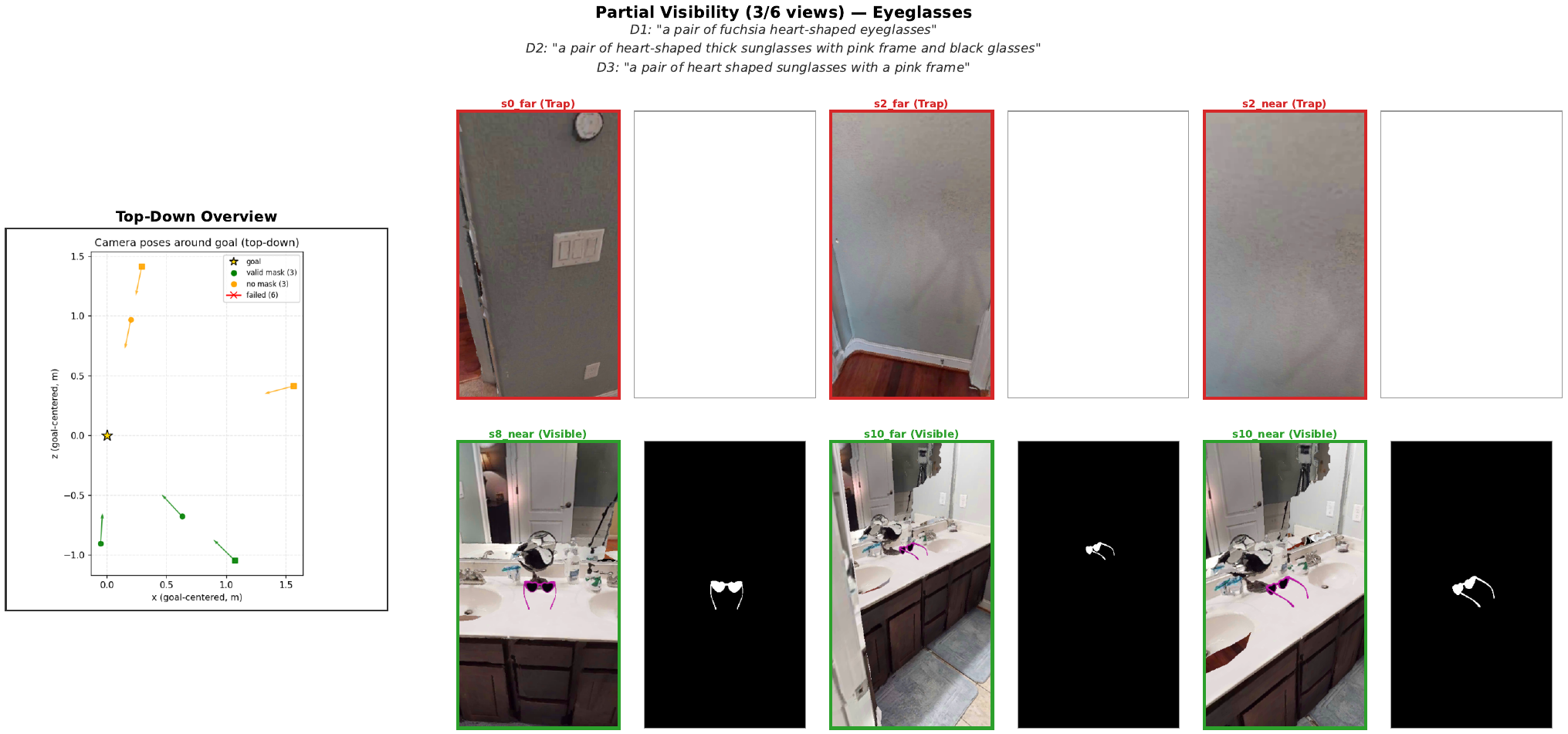}
  \caption{\textbf{Multi-view capture with trap views.} Red borders indicate trap views (\texttt{mask\_meets\_threshold} = false): geometrically navigable but semantically uninformative. Green borders mark sectors with valid visibility.}
  \label{fig:capture_traps}
\end{figure}

Figures~\ref{fig:capture_full}--\ref{fig:capture_traps} show two concrete episodes that complement the topology diagram in the main paper (Fig.~\ref{fig:sector_topology}): a full-visibility episode where every navigable sector observes the target, and a partial-visibility episode where several sectors yield trap views (the target object's segmentation mask falls below the visibility threshold).

\section{Prompt Templates}
\label{app:prompts}

Below are the prompt templates used by the reference agent. Placeholders in \texttt{\{curly braces\}} are filled at runtime with episode-specific data.

\newtcolorbox{promptbox}[1][]{
  enhanced,
  colback=gray!3,
  colframe=black!50,
  fonttitle=\bfseries\small\sffamily,
  title={#1},
  breakable,
  left=5pt, right=5pt, top=3pt, bottom=3pt,
  fontupper=\ttfamily\footnotesize,
  boxrule=0.5pt,
  arc=2pt
}

\subsection{Category Classification}
\label{app:prompt_category}

A text-only prompt that maps the three descriptions to one of the 18 predefined categories.

\begin{promptbox}[Category Classification]
You will classify the OBJECT CATEGORY described by three
sentences into EXACTLY ONE of the following classes:
[backpack, bag, ball, book, camera, cellphone, eyeglasses,
hat, headphones, keys, laptop, mug, shoes, teddy bear,
toy, visor, wallet, watch]

Descriptions:
1) \{desc1\}
2) \{desc2\}
3) \{desc3\}

RULES:
- Output only the chosen class name, exactly as listed
  (case-insensitive allowed).
- Do NOT output any extra words, explanations, punctuation,
  or quotes.
- If multiple possible, choose the most specific one among
  the list.

Answer with the single class name only.
\end{promptbox}

\subsection{Description Merging}
\label{app:prompt_merge}

The merge prompt is used in the single-view Merged mode to combine three descriptions into a single sentence before holistic verification.

\begin{promptbox}[Description Merge]
You are a description merger for object verification.

Given three descriptions of the SAME object, combine them
into ONE concise sentence.

CRITICAL RULES:
1. PRESERVE distinctive features that help distinguish this
   object from similar ones
   - Keep: specific colors, patterns, logos, text, brand
     names, unique marks
   - Keep: distinguishing parts (e.g., "broken zipper",
     "red bow", "Apple logo")
2. REMOVE redundant or generic information
   - Remove: repeated facts mentioned in multiple
     descriptions
   - Remove: vague descriptions like "nice" or "common"
3. The merged description should be SPECIFIC enough to
   identify THIS exact object

Description 1: \{desc1\}
Description 2: \{desc2\}
Description 3: \{desc3\}

Output only the merged description in one sentence,
no explanation.
\end{promptbox}

\subsection{Attribute Extraction}
\label{app:prompt_extract}

The attribute extraction prompt decomposes three descriptions into a structured list of verifiable attributes.

\subsubsection{Standard Attribute Extraction}

\begin{promptbox}[Attribute Extraction]
You are an attribute extractor for instance verification.

INPUT
- target class: "\{class\_text\}"
- natural descriptions:
  1) "\{desc1\}"
  2) "\{desc2\}"
  3) "\{desc3\}"

TASK
Extract visual attributes for verifying this object
instance.

For each attribute:
- name: Include the PART if the attribute belongs to a
  sub-part
  - If describing the main object: use general name
    (e.g., "color", "shape")
  - If describing a sub-part: use "part\_attribute"
    format (e.g., "scarf\_color", "nose\_color",
    "bow\_pattern")
- type: \{color, pattern, part, logo\_text, material,
  relation, count, shape, other\}
- weight: \{1,2,3\} (3=most discriminative)
- evidence\_phrase: EXACT value from description

CRITICAL RULES:
1. ONLY extract EXPLICITLY mentioned attributes
2. NEVER invent values not in the descriptions
   (NO hallucination!)
3. For sub-parts (scarf, nose, bow), include part name:
   "scarf\_color", "bow\_type"
4. evidence\_phrase must come from the actual text

EXAMPLE for teddy bear:
"a light blue teddy bear with a transparent white scarf
with red dots"

CORRECT:
\{"name": "color", "evidence\_phrase": "light blue"\}
  <- main object color
\{"name": "scarf\_color",
  "evidence\_phrase": "transparent white"\}
  <- scarf's color
\{"name": "scarf\_pattern",
  "evidence\_phrase": "red dots"\}
  <- scarf's pattern

WRONG:
\{"name": "material",
  "evidence\_phrase": "transparent white"\}
  <- loses context!

RETURN JSON ONLY (4\textasciitilde\{max\_attrs\} attributes):
\{
  "attributes":[
    \{"name":"...", "type":"...", "weight":2,
     "evidence\_phrase":"..."\}
  ]
\}
\end{promptbox}

\noindent At runtime, \texttt{\{max\_attrs\}} is set to $N_{\max}{=}8$ (\S\ref{sec:agent}); we ask for at least 4 attributes to maintain coverage when descriptions are sparse.

\subsection{Attribute Verification}
\label{app:prompt_verify}

The verification prompt is a vision-language call that takes a cropped image of the candidate object and checks one attribute at a time. Two variants exist: standard and direct.

\subsubsection{Standard Attribute Verification}

\begin{promptbox}[Attribute Verification]
You will be given an image of a candidate object.
We need to verify BOTH the object category AND a specific
attribute.

IMPORTANT: The object must be the correct CATEGORY first.
If it's not a \{object\_category\}, answer 'No' immediately.

Object Category (MUST match): \{object\_category\}

Object descriptions (3 sentences describing the SAME
object):
1) \{desc1\}
2) \{desc2\}
3) \{desc3\}

Attribute to verify:
\{attr\_name\} = \{expected\_value\}

Answer in JSON format:
\{"answer": "Yes/No/Unsure", "reason": "brief
explanation"\}

Rules:
- FIRST check: Is the object in the image a
  \{object\_category\}? If NOT, say 'No'.
- If it IS the correct category, then check the attribute.
- If the attribute matches the expected value, say 'Yes'.
- If the attribute is visible but has a DIFFERENT value
  (e.g. red instead of blue), say 'No'.
- If the attribute is NOT visible due to angle/occlusion,
  say 'Unsure'.
\end{promptbox}

\subsubsection{Direct Verification}

In direct mode, each description is verified holistically rather than through individual attributes.

\begin{promptbox}[Direct Verification]
You will be given an image of a candidate object.
Task: Does this image show the described target item?

IMPORTANT: The object must be a \{object\_category\}.
If it's a different type of object, answer 'No'.

Object Category (MUST match): \{object\_category\}

Target description:
\{desc\}

Answer in JSON format:
\{"answer": "Yes/No/Unsure", "reason": "brief
explanation"\}

Rules:
- FIRST check: Is the object in the image a
  \{object\_category\}? If NOT, say 'No'.
- If the category matches AND the object matches the
  description, say 'Yes'.
- If the category matches but description doesn't match
  (e.g. wrong color/type), say 'No'.
- If the object is not visible, occluded, or the image
  is unclear, say 'Unsure'.
\end{promptbox}

\subsection{Navigation Decision}
\label{app:prompt_nav}

The navigation prompt guides the MLLM to select the next viewing direction based on the current verification state. Two variants correspond to the two multi-view query modes (attribute decomposition and direct description); the merged mode uses only single-view evaluation and does not require a navigation prompt.

\subsubsection{Attribute-Based Navigation}

\begin{promptbox}[Attribute-Based Navigation]
You are an embodied perception planner for attribute-based
object verification.

Task: The following 3 sentences ALL describe the SAME target
object. We have decomposed these descriptions into specific
attributes to verify.

Object Description (3 sentences about the SAME object):
1) \{desc1\}
2) \{desc2\}
3) \{desc3\}

Object Category: \{object\_category\}

Attribute Verification Status:
\{verification\_status\}

Navigation State:
- Current View: You are currently viewing the object from
  the FRONT
- Already Visited: \{visited\_directions\}
- Available Directions to Move: \{allowed\_directions\}

\{visibility\_warning\}

Direction Guide (all directions are relative to YOUR
current position):
\{direction\_guide\}

IMPORTANT: You MUST choose ONE direction STRICTLY from the
Available Directions to Move list above.
- Do NOT choose a direction that is NOT in the list.
- Do NOT choose a direction listed as "Already Visited".

Your Task: Analyze each available direction, then choose
the best one.

STEP 1 -- Per-Direction Analysis (you MUST analyze ALL
available directions):
For EACH direction in \{allowed\_directions\}, assess:
a) Reachability: Look at the scene around the object.
   Is the path in that direction blocked by walls,
   furniture, or other obstacles?
b) Target Visibility: If you move there, will you be
   able to see the target object, or might it be occluded?
c) Information Gain: Which missing/unverified attributes
   could you potentially observe from that angle?

STEP 2 -- Direction Decision:
Based on your Step 1 analysis, choose the direction that:
1. Is most likely reachable (clear path, no obstacles)
2. Will likely give you a clear view of the target object
3. Has the highest potential to reveal missing/unverified
   attributes

Output EXACTLY this JSON format:
\{"per\_direction\_analysis": "<your analysis of each
 available direction>", "chosen\_direction": "<must be
 from Available Directions>", "why": "<brief reason
 linking to missing attributes and reachability>"\}
\end{promptbox}

\subsubsection{Direct Description Navigation}

The direct-mode variant uses the same prompt structure but reports per-description match confidence instead of per-attribute states, and assesses information gain over unverified descriptions rather than attributes.

\begin{promptbox}[Direct Navigation -- key differences]
Current Verification Status:
\{verification\_status\}

(Verification status shows per-description match
confidence rather than per-attribute states.)

STEP 1c) Information Gain: Which unverified aspects of
the description could you potentially observe from that
angle?

STEP 2.3: Has the highest potential to reveal unverified
description aspects
\end{promptbox}


\section{Evaluation Index Generation}
\label{app:index_generation}

The evaluation index is built from raw capture data in two stages: index scanning and pair construction.

\subsection{Raw Index Building}
\label{app:raw_index}

The first stage scans all captured episode directories and builds a raw index of available episodes.

\paragraph{Input.}
Each episode is stored as a directory containing a \texttt{meta.json} file with the following fields:
\begin{itemize}[leftmargin=2em]
  \item \texttt{object\_id}: unique identifier of the target object in the scene.
  \item \texttt{object\_category}: one of the 18 category labels.
  \item \texttt{scene\_key}: HM3D scene identifier.
  \item \texttt{episode\_id}: unique episode identifier.
  \item \texttt{viewpoints} (or \texttt{captures} in v1 format): an array of viewpoint records, each containing:
    \begin{itemize}
      \item \texttt{navigable}: whether a valid NavMesh position was found.
      \item \texttt{mask\_meets\_threshold}: whether the object's semantic mask exceeds the category-specific threshold (Table~\ref{tab:mask_thresholds}).
      \item \texttt{sector\_index}: the active sector index, drawn from the raw set $\{0, 2, 4, 6, 8, 10\}$.
      \item \texttt{rgb}: filename of the captured RGB image.
    \end{itemize}
\end{itemize}

\paragraph{Process.}
The index builder iterates over all episode metadata files matching the pattern below. For each episode, it:

\smallskip
\noindent\texttt{\{dataset\_root\}/\{split\}/**/meta.json}
\smallskip
\begin{enumerate}[leftmargin=2em]
  \item Parses the viewpoint array (supporting both v1 and v2 format).
  \item Identifies valid start sectors: sectors where at least one viewpoint has \texttt{mask\_meets\_threshold = true}.
  \item Counts navigable sectors (sectors with at least one navigable viewpoint) and mask-visible sectors.
  \item Emits a JSONL record with episode metadata and sector statistics.
\end{enumerate}

\paragraph{Output.}
A temporary raw index file (\texttt{\_tmp\_raw.jsonl}) where each line is a JSON object:
\begin{verbatim}
{
  "episode_path": "val/scene_name/episode_id",
  "scene": "scene_key",
  "episode": "episode_id",
  "target_object_id": "object_id",
  "target_object_category": "category",
  "valid_start_sectors": [0, 2, 4],
  "navigable_sectors": [0, 2, 4, 6, 8, 10],
  "n_navigable": 6,
  "n_mask_visible": 5
}
\end{verbatim}

\subsection{Pair Construction}
\label{app:negative_sampling}

For each target object we sample one positive pair (\texttt{query\_object\_id} = \texttt{target\_object\_id}, \texttt{label}=1), one neg\_same pair (uniform random sample from same-category distractor pool, capped at 10 distractors per category), and one neg\_diff pair (uniform random sample from cross-category pool, capped at 2 per other category). neg\_same query descriptions come from the distractor while the scene still contains the original target. We do not impose visual-similarity constraints on neg\_diff distractors; difficulty-stratified selection is left to future benchmark versions. The standard evaluation set contains $N = 3{,}000$ pairs balanced 1{:}1{:}1.


\section{Capture Dataset Statistics}
\label{app:capture_stats}

We report capture statistics for both training and validation splits of \bench{}, produced by the pipeline in App.~\ref{app:capture_pipeline}.

\subsection{Object Instance Gallery}

\begin{figure}[htbp]
\centering
\includegraphics[width=\textwidth,height=0.85\textheight,keepaspectratio]{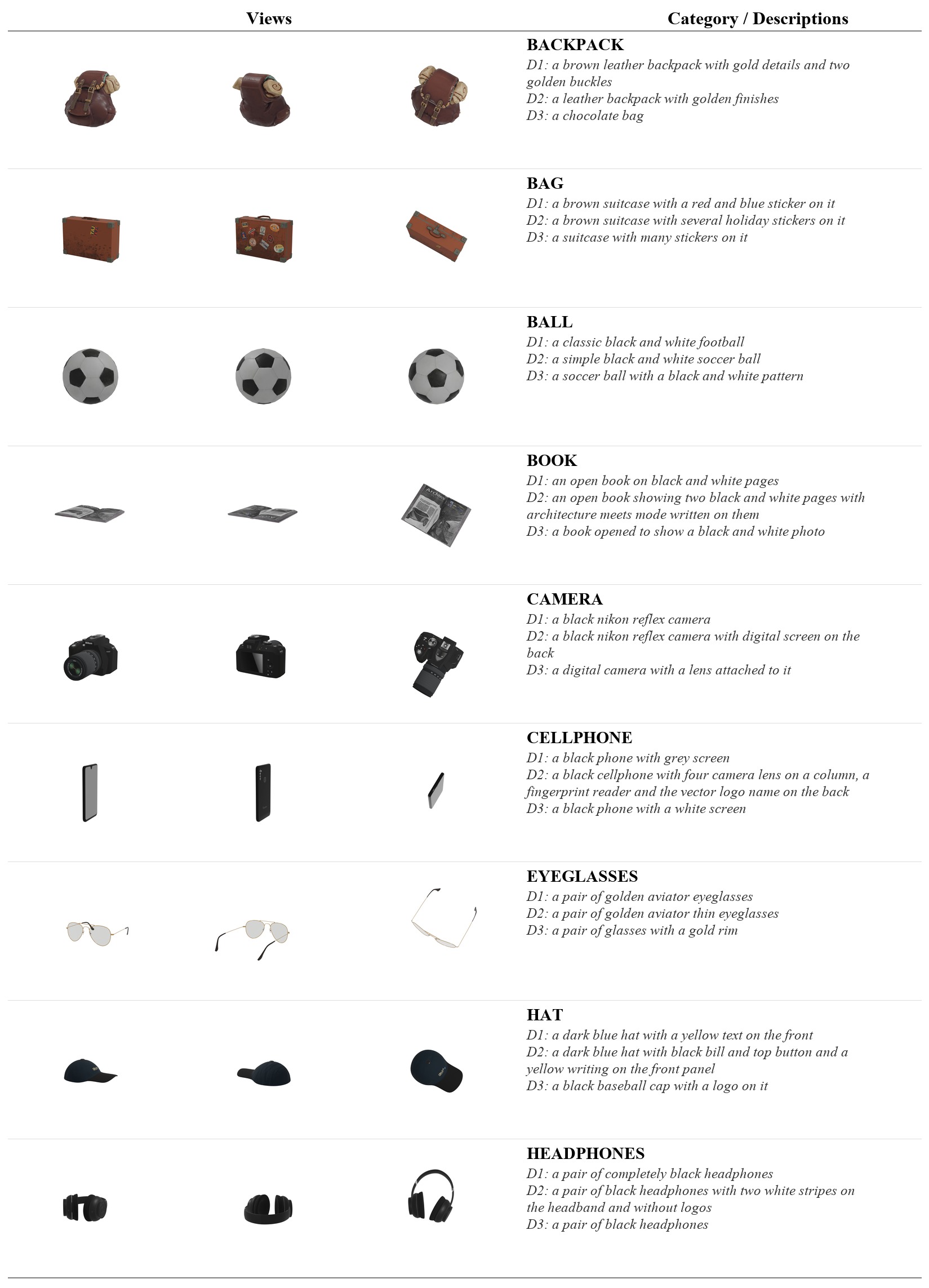}
\caption{Representative \bench{} instances (categories backpack--headphones). Each row shows one object across three ground-truth viewpoints alongside its three descriptions D1--D3.}
\label{fig:app_gallery_1}
\end{figure}

\begin{figure}[htbp]
\centering
\includegraphics[width=\textwidth,height=0.85\textheight,keepaspectratio]{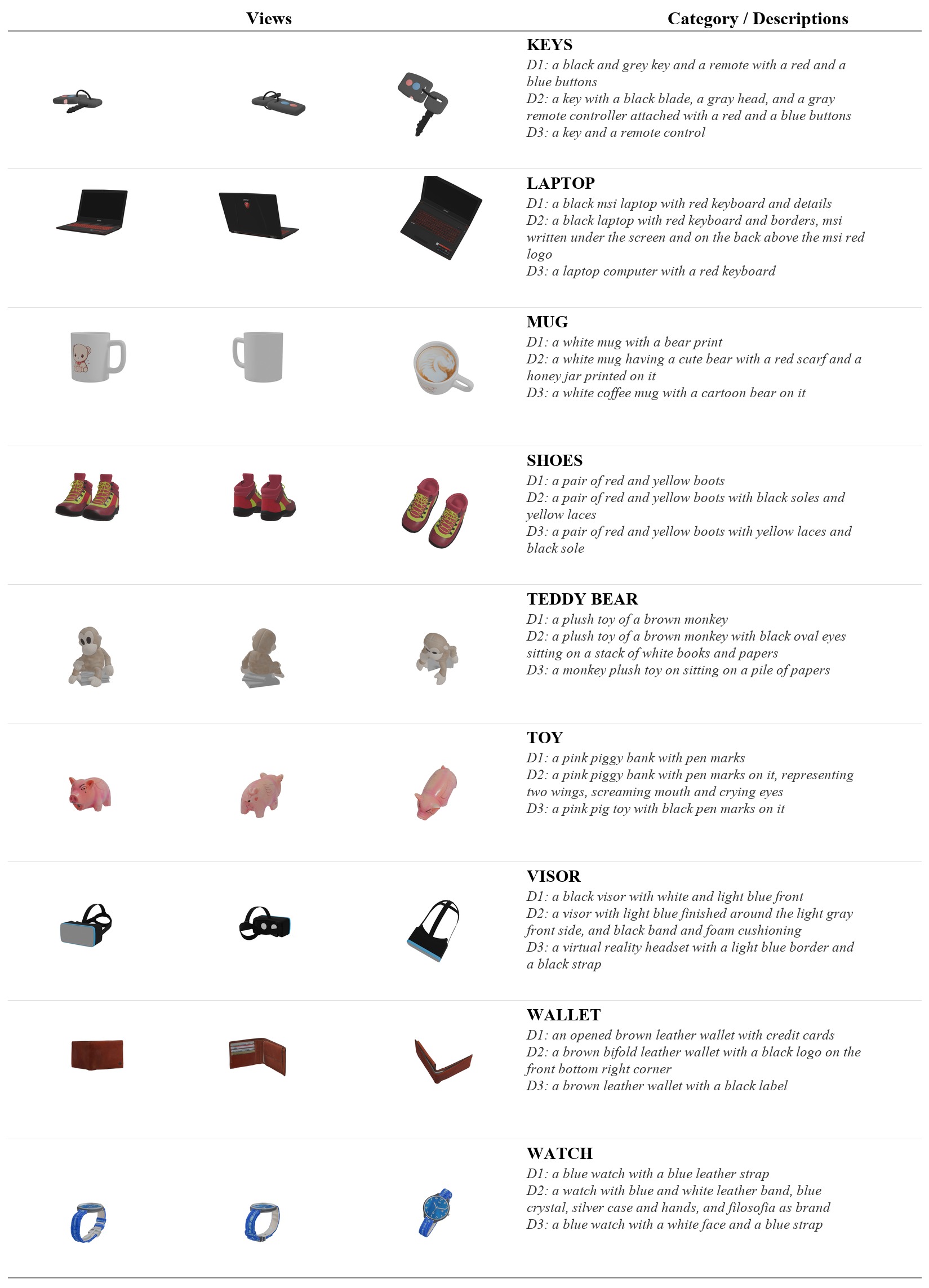}
\caption{Representative \bench{} instances (categories keys--watch). Spans 18 everyday categories of varying visual complexity.}
\label{fig:app_gallery_2}
\end{figure}

Figures~\ref{fig:app_gallery_1} and~\ref{fig:app_gallery_2} show one representative instance per object category, with three ground-truth viewpoints and the three independently written natural-language descriptions used as the verification query. Description specificity ranges from concise category-level identifiers (e.g., ``a chocolate bag'') to fine-grained attribute-level characterizations (e.g., ``a brown leather backpack with gold details and two golden buckles''); we preserve this variation rather than normalizing it.

\subsection{Overall Summary}

Table~\ref{tab:app_capture_overview} compares the two splits at a high level.

\begin{table}[htbp]
\centering
\caption{PInVerify capture dataset statistics.}
\label{tab:app_capture_overview}
\begin{tabular}{l r r}
\toprule
\textbf{Property} & \textbf{Train} & \textbf{Val} \\
\midrule
Scenes & 145 & 35 \\
Episodes & 37,583 & 2,485 \\
Unique objects & 264 & 71 \\
Object categories & 18 & 18 \\
\midrule
Avg navigable sectors / ep & 4.65 & 4.66 \\
Avg visible sectors / ep & 3.81 & 3.88 \\
Avg trap sectors / ep & 0.84 & 0.78 \\
\midrule
Avg navigable viewpoints / ep & 7.87 & 7.82 \\
Avg valid-mask viewpoints / ep & 6.46 & 6.49 \\
\midrule
Median mask area (px) & 2,783 & 3,113 \\
\bottomrule
\end{tabular}
\end{table}

\subsection{Per-Category Distribution}

\begin{figure}[htbp]
\centering
\includegraphics[width=\textwidth]{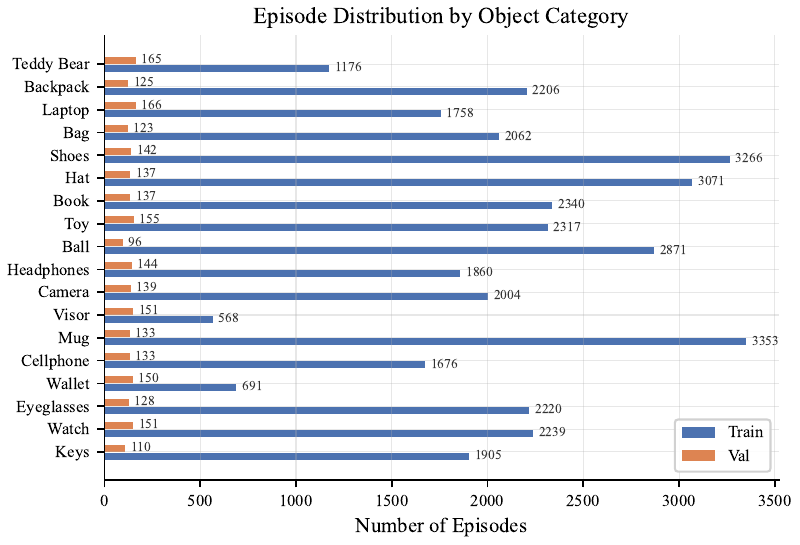}
\caption[Episode distribution by object category]{Episode distribution by object category for the training and validation splits. Categories are ordered by typical object size (large to small). The training set contains approximately 15$\times$ more episodes than the validation set per category, with proportional scaling.}
\label{fig:app_category_dist}
\end{figure}

Figure~\ref{fig:app_category_dist} shows the episode count for each of the 18 object categories, comparing the training and validation splits. The number of episodes per category reflects the number of distinct object instances available in the PInNED source dataset and the number of HM3D scenes into which they can be injected.

Table~\ref{tab:app_per_category} breaks down the training split by category: number of unique instances, average sector navigability, and median mask area.

\begin{table}[htbp]
\centering
\caption{Per-category statistics of the capture dataset (train split).}
\label{tab:app_per_category}
\setlength{\tabcolsep}{3.5pt}
\small
\begin{tabular}{l r r r r r r}
\toprule
\textbf{Category} & \textbf{Eps} & \textbf{Objs} & \textbf{Nav} & \textbf{Vis} & \textbf{Trap} & \textbf{Mask (med)} \\
\midrule
    Teddy Bear & 1,176 & 8 & 4.6 & 4.0 & 0.7 & 10,042 \\
    Backpack & 2,206 & 15 & 4.7 & 3.9 & 0.7 & 10,583 \\
    Laptop & 1,758 & 12 & 4.6 & 3.9 & 0.7 & 8,740 \\
    Bag & 2,062 & 14 & 4.7 & 3.9 & 0.7 & 5,949 \\
    Shoes & 3,266 & 22 & 4.7 & 3.9 & 0.8 & 3,992 \\
    Hat & 3,071 & 21 & 4.6 & 3.9 & 0.7 & 5,183 \\
    Book & 2,340 & 16 & 4.6 & 3.9 & 0.8 & 5,224 \\
    Toy & 2,317 & 16 & 4.7 & 3.8 & 0.8 & 2,090 \\
    Ball & 2,871 & 22 & 4.6 & 3.8 & 0.8 & 3,764 \\
    Headphones & 1,860 & 13 & 4.7 & 3.9 & 0.8 & 2,995 \\
    Camera & 2,004 & 14 & 4.7 & 3.8 & 0.9 & 2,444 \\
    Visor & 568 & 4 & 4.6 & 3.9 & 0.8 & 3,090 \\
    Mug & 3,353 & 23 & 4.7 & 3.7 & 0.9 & 1,289 \\
    Cellphone & 1,676 & 12 & 4.7 & 3.7 & 0.9 & 686 \\
    Wallet & 691 & 5 & 4.6 & 3.8 & 0.9 & 840 \\
    Eyeglasses & 2,220 & 16 & 4.6 & 3.6 & 1.1 & 445 \\
    Watch & 2,239 & 16 & 4.7 & 3.7 & 1.0 & 317 \\
    Keys & 1,905 & 15 & 4.7 & 3.6 & 1.1 & 168 \\
\midrule
    \textbf{Total / Avg} & 37,583 & 264 & 4.7 & 3.8 & 0.8 & 2,783 \\
\bottomrule
\end{tabular}
\end{table}

\subsection{Sector and Viewpoint Statistics}

Up to 12 viewpoints per episode (6 sectors $\times$ 2 distance levels) are sampled. Not all sectors are navigable, and navigable sectors may still be trap views.

\begin{figure}[htbp]
\centering
\includegraphics[width=\textwidth]{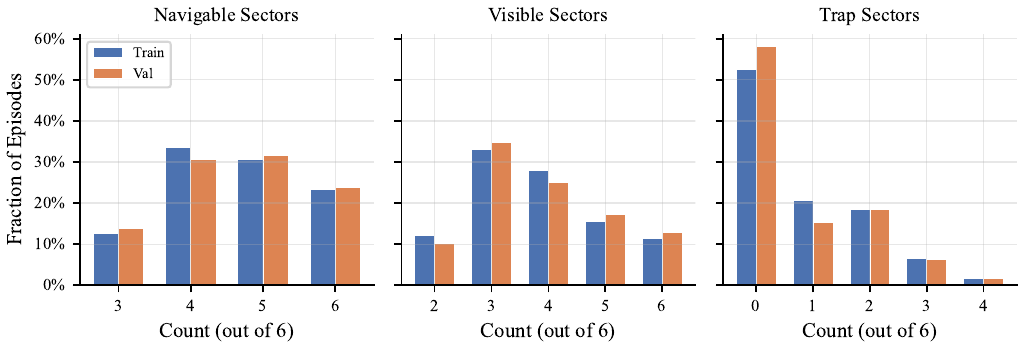}
\caption[Distribution of navigable, visible, and trap sectors per episode]{Distribution of navigable, visible, and trap sectors per episode (out of 6 total sectors). Percentages are normalized within each split. \textbf{Left}: Navigable sectors (agent can reach the sector). \textbf{Center}: Visible sectors (object mask exceeds category threshold). \textbf{Right}: Trap sectors (navigable but object not meaningfully visible).}
\label{fig:app_sector_dist}
\end{figure}

Figure~\ref{fig:app_sector_dist} shows the per-episode distribution of navigable, visible, and trap sectors. Most episodes have 4--5 navigable sectors, 3--4 visible sectors, and 0--1 trap sectors. The two splits have consistent distributions.

\subsection{Per-Category Sector Quality}

\begin{figure}[htbp]
\centering
\includegraphics[width=\textwidth]{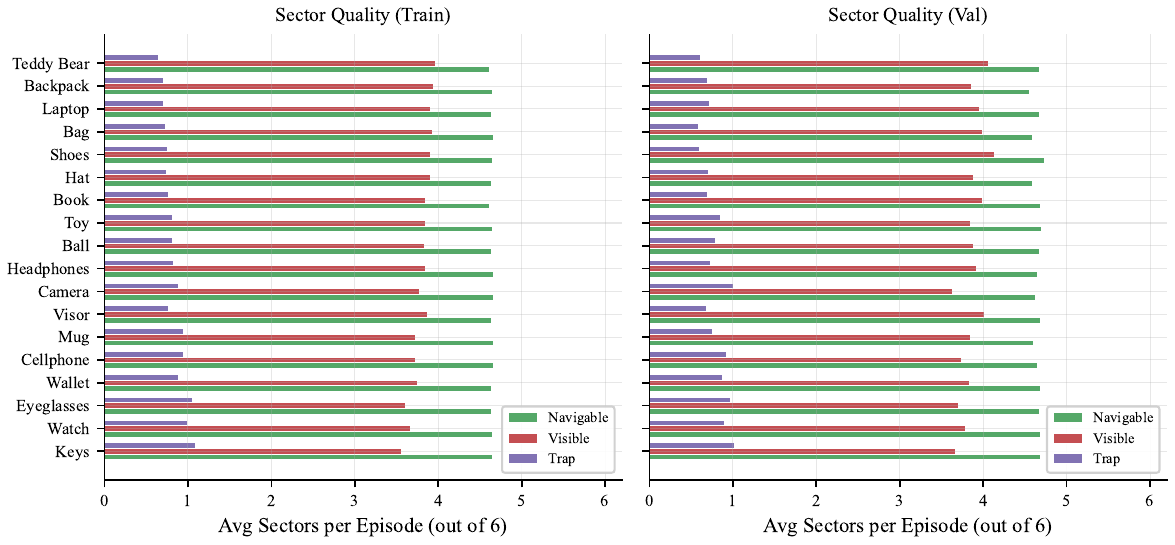}
\caption[Per-category sector quality]{Per-category average number of navigable, visible, and trap sectors per episode (training split). Categories are ordered by typical object size. Small objects such as keys, eyeglasses, and watch have the highest trap-sector rates.}
\label{fig:app_cat_sector}
\end{figure}

Figure~\ref{fig:app_cat_sector} reports per-category averages of navigable, visible, and trap sector counts. Small objects (keys, watch, eyeglasses) have higher trap-sector rates because their projected mask falls below the visibility threshold from many angles. Large objects (teddy bear, backpack, laptop) maintain high visibility across navigable sectors.

\subsection{Mask Area Distribution}

\begin{figure}[htbp]
\centering
\includegraphics[width=0.85\textwidth]{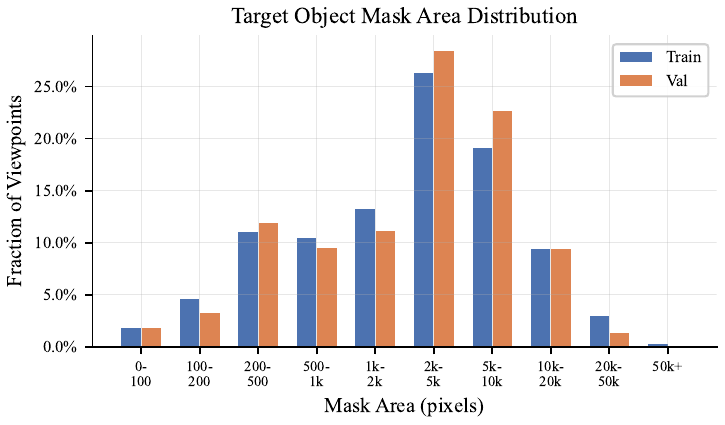}
\caption[Mask area distribution]{Distribution of target object mask area across all navigable viewpoints with detected masks. The training and validation splits exhibit similar distributions, peaking in the 2k--5k pixel range.}
\label{fig:app_mask_hist}
\end{figure}

The target object's mask area (in pixels) quantifies how prominently the object appears in each viewpoint's observation. Figure~\ref{fig:app_mask_hist} compares the overall mask area distribution between the training and validation splits.

\begin{figure}[htbp]
\centering
\includegraphics[width=0.85\textwidth]{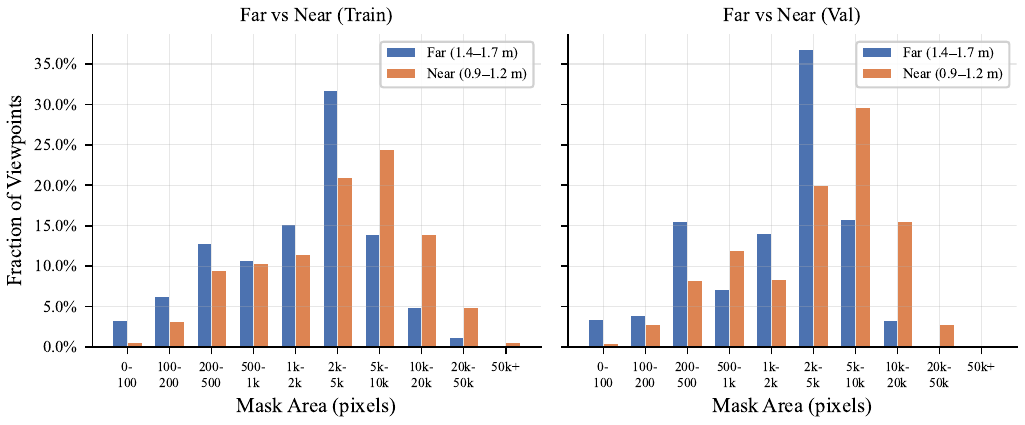}
\caption[Mask area distribution: far vs.\ near viewpoints]{Mask area distribution for far vs.\ near viewpoints (training split). Near views produce systematically larger masks, providing finer visual detail for attribute verification.}
\label{fig:app_mask_far_near}
\end{figure}

Figure~\ref{fig:app_mask_far_near} compares mask areas between far (1.4--1.7\,m) and near (0.9--1.2\,m) viewpoints. As expected, near viewpoints produce larger mask areas due to the closer camera distance, with the distribution shifted toward higher pixel counts.

\begin{figure}[htbp]
\centering
\includegraphics[width=\textwidth]{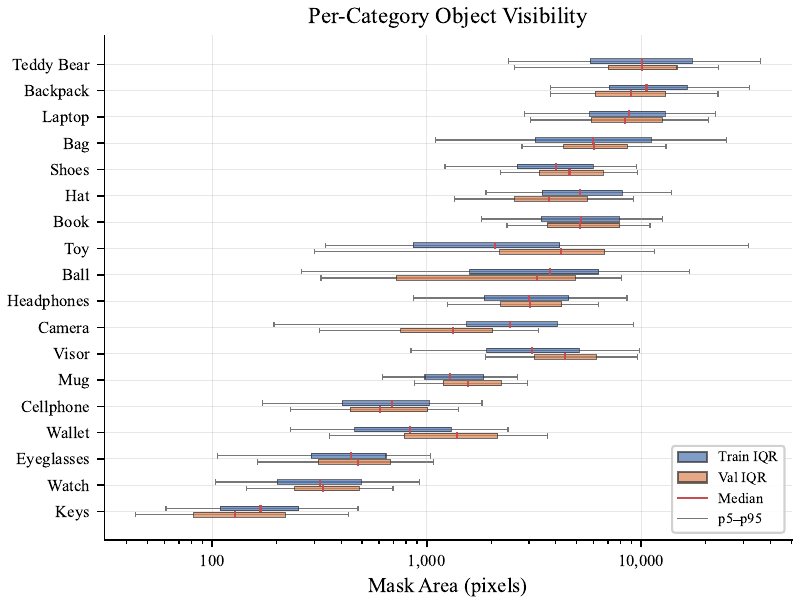}
\caption[Per-category mask area distribution]{Per-category mask area distribution (training split). Boxes show the interquartile range (p25--p75), red lines indicate the median, and whiskers extend to the 5th and 95th percentiles. The $x$-axis uses a logarithmic scale.}
\label{fig:app_cat_mask}
\end{figure}

Figure~\ref{fig:app_cat_mask} presents per-category mask area statistics using a box-plot representation. The ordering from large-mask categories (teddy bear, backpack) to small-mask categories (watch, keys) spans nearly two orders of magnitude, illustrating large variation in visual prominence across object types.

\section{Training Pool Construction}
\label{app:training_pool}

We sample from the training scenes (145 HM3D scenes, 37{,}583 raw episodes) two disjoint episode pools. The \emph{SFT pool} contains 5{,}075 episodes (35 per scene, used for supervised fine-tuning), and the \emph{RL pool} contains another 5{,}075 episodes (used for GRPO / GSPO reinforcement learning). Within each scene, episodes are deduplicated by (rounded position at 0.1\,m tolerance, object\_id). The same position may appear in both pools but with different target objects, so no visual observation is shared between SFT and RL training. A deterministic per-scene seed ensures reproducibility. Each pool generates its own pair index using the same negative sampling procedure as the evaluation set: for each target object, a positive pair, a neg\_same pair (same-category distractor), and a neg\_diff pair (cross-category distractor) are sampled, maintaining a balanced 1{:}1{:}1 distribution.

\section{Training Details}
\label{app:training_details}

All training is performed on 4$\times$NVIDIA RTX~3090 using the ms-swift framework~\cite{zhao2025swift}. We apply LoRA (rank 16, $\alpha = 32$) to all linear layers of the language model; the vision encoder is frozen.
For all DINO-based evaluation/inference results, Grounding DINO uses box threshold 0.25 and text threshold 0.25. Returned boxes are cropped with a 3-pixel padding and bicubically upsampled to a 512\,px minimum side, matching the reference-agent implementation.

\textbf{SFT.} The model is trained on $\sim$22K samples from the SFT pool (App.~\ref{app:training_pool}). Each sample is a multi-step trajectory annotated with structured chain-of-thought labels templated directly from ground-truth pair information and per-step visibility metadata supplied by the benchmark (full schema in App.~\ref{app:sft_format}). Standard hyperparameters: learning rate $1\times 10^{-4}$ with cosine schedule, warmup ratio 0.05, max length 2048, per-device batch size 2 with gradient accumulation 4 (effective batch size 32), bf16 mixed precision, gradient checkpointing, and 3 epochs with epoch-level checkpointing.

\textbf{Post-SFT alignment.} Starting from the SFT adapter we explore three alignment strategies. \emph{DPO}~\cite{rafailov2024direct} uses preference pairs constructed by perturbing SFT targets, with $\beta = 0.1$, max length 2048, per-device batch size 1 with gradient accumulation 4, warmup ratio 0.05, and learning rate $5\times 10^{-7}$. We evaluate two checkpoints: DPO-200 (200 training steps, where the reward margin approximately converges) and DPO-400 (400 steps, to verify stability beyond convergence). \emph{GRPO}~\cite{shao2024deepseekmath} generates $G{=}4$ completions per prompt with temperature 1.0, max completion length 1024, per-device batch size 1 with gradient accumulation 4, warmup ratio 0.05, learning rate $1\times 10^{-6}$, and 1 epoch on the RL pool. \emph{GSPO}~\cite{zheng2025gspo} keeps the same hyperparameters but replaces GRPO's token-level importance ratio with a length-normalized sequence-level ratio.

\textbf{Reward.} GRPO/GSPO use the implementation in \texttt{training/reward.py}. The combined reward is
\[
r = 0.5\,r_{\text{verify}} + 0.4\,r_{\text{action}} + 0.1\,r_{\text{format}}.
\]
For verification, exact label match scores 1.0; partial credit is 0.4 for \texttt{Unsure}$\rightarrow$\texttt{Yes}, 0.2 for \texttt{Unsure}$\rightarrow$\texttt{No}, and 0.3 for overconfident \texttt{Yes}/\texttt{No} when the ground truth is \texttt{Unsure}; \texttt{Yes}$\leftrightarrow$\texttt{No} scores 0.0. For actions, correct \texttt{STOP} scores 1.0; premature stop and over-exploration both score 0.1; \texttt{MOVE} to a best sector / other visible+navigable sector / navigable-only sector / unreachable sector scores 1.0 / 0.7 / 0.3 / 0.0; a \texttt{MOVE} with no valid direction string scores 0.3. For format, responses with both \texttt{<think>} and \texttt{<answer>} score 1.0, those with only \texttt{<answer>} score 0.5, and malformed outputs score 0.0.

\section{Scope, Limitations, and Release Notes}
\label{app:scope_release}

\bench{} reuses scenes, objects, and descriptions from PInNED~\cite{barsellotti2024personalized}; the contribution is the task formulation, capture pipeline, and protocol-level annotations rather than new visual data, so we inherit that substrate's biases (indoor residential HM3D scenes, Objaverse-XL objects with occasionally incongruent placements, and English-only descriptions). The evaluation is offline and does not test live-control challenges such as lighting drift, dynamic objects, or compute budgets. Compute constraints prevented multi-seed runs; we therefore report 95\% binomial CIs and recommend paired bootstrap or McNemar tests for close comparisons. Cross-model conclusions are intentionally limited to open-source on-device backbones at $\leq$8B parameters. All trained agents apply LoRA only to the language model, and the training-free pipeline caps attribute decomposition at $N_{\max}{=}8$.

We plan to release the \bench{} dataset, evaluation environment, prompts, training data, the training-free reference baseline, and the LoRA fine-tuning pipeline (SFT + DPO / GRPO / GSPO) under a permissive license; trained checkpoints are planned to accompany the release. The release is hosted at \url{https://github.com/Avalon-S/PInVerify}.

\section{Dataset Format Specification}
\label{app:data_format}

We document the on-disk formats: capture metadata, eval/train indices, and SFT/RL training samples.

\subsection{Directory Structure}

The PInVerify dataset is organized as follows:

\begin{verbatim}
pv_dataset/
  pin_capture/                  # Sector-based capture data
    train/                      # Training scenes
      {scene_id}/{episode}/     # Per-episode directory
        meta.json               # Episode metadata
        overview.png            # Top-down camera pose visualization
        rgb/                    # RGB images (rgb_s0_far.png, ...)
        mask/                   # Segmentation masks
    val/                        # Validation scenes (same structure)
  val/                          # Evaluation indices
    pv_index_50.jsonl           # 50-episode subset (smoke test)
    pv_index_100.jsonl          # 100-episode subset
    pv_index_500.jsonl          # 500-episode subset
    pv_index_1000.jsonl         # 1000-episode subset
    pv_index_all.jsonl          # Full evaluation set (3,000 episodes)
    pv_index_all_7455.jsonl     # Extended set (7,455 episodes)
    object_goal_distractors_map.json   # Per-target distractor pool
  train_sft/                    # SFT training data
    pv_train_sft_index.jsonl    # Pair index (15,225 entries)
    sft_data.jsonl              # SFT samples (~22K)
    crops/                      # GT bbox crops for SFT
    object_goal_distractors_map.json
  train_rl/                     # RL training data
    pv_train_rl_index.jsonl     # Pair index (15,225 entries)
    rl_data.jsonl               # RL prompts (~19K)
    dpo_data.jsonl              # DPO preference pairs
    crops_rl/                   # GT bbox crops for RL
    crops_dpo/                  # GT bbox crops for DPO
    object_goal_distractors_map.json
  image_gt/                     # Reference object images (by category)
    {category}/                 # 18 category subdirectories
      {object_id}_0.png        # 3 views per object
      {object_id}_1.png
      {object_id}_2.png
  object_descriptions.json      # Object ID -> 3 human descriptions
  object_descriptions_with_category.json  # Descriptions + category
  attr_cache.json               # Object ID -> extracted attributes
  category_cache.json           # Object ID -> predicted category
  merge_cache.json              # Merged description cache
\end{verbatim}

\subsection{Capture Metadata (\texttt{meta.json})}
\label{app:meta_json}

Each episode directory contains a \texttt{meta.json} file that describes the scene, target object, camera configuration, and all captured viewpoints. Listing~\ref{lst:meta_json} shows the top-level structure (viewpoint array truncated for brevity).

\begin{lstlisting}[
  caption={Capture metadata structure (\texttt{meta.json}). Only the first viewpoint is shown; each episode contains up to 12 viewpoints (6 sectors $\times$ 2 ranges).},
  label={lst:meta_json},
  basicstyle=\ttfamily\scriptsize,
  frame=single,
  breaklines=true,
  numbers=left,
  numberstyle=\tiny\color{gray},
  xleftmargin=2em
]
{
  "scene_id": "data/scene_datasets/hm3d/train/
               00059-kJxT5qssH4H/kJxT5qssH4H.basis.glb",
  "scene_key": "00059-kJxT5qssH4H",
  "episode_id": "0",
  "goal_position_nominal": [2.612, 3.301, -1.106],
  "object_category": "backpack",
  "object_id": "2ad86321197a49feb54b7726743d7fd0",
  "camera_intrinsics": {
    "width": 360, "height": 640,
    "hfov_deg": 42.0, "vfov_deg": 68.62
  },
  "sensor_mount_offset_y": 1.31,
  "n_sectors": 12,
  "sector_order": [0, 2, 4, 6, 8, 10],
  "sector_skip": 1,
  "ranges": {"near": [0.9, 1.2], "far": [1.4, 1.7]},
  "category_mask_threshold": 500,
  "viewpoints": [
    {
      "tag": "s0_far",
      "navigable": true,
      "in_frustum": true,
      "has_mask": true,
      "mask_area_px": 5111,
      "mask_bbox_xyxy": [140, 368, 222, 452],
      "mask_area_fraction": 0.0222,
      "mask_meets_threshold": true,
      "rgb": "rgb/rgb_s0_far.png",
      "mask_raw_path": "mask/mask_s0_far.png",
      "camera_position": [4.261, 4.771, -1.039],
      "camera_rotation_quat_wxyz": [0.697, -0.187, 0.669, 0.179],
      "sector_index": 0,
      "angle_deg_range": [0, 30],
      "range_label": "far",
      "radius_m": 1.65
    },
    ... // 11 more viewpoints (s0_near, s2_far, s2_near, ...)
  ],
  "navigable_count": 6,
  "valid_mask_count": 6,
  "success": true
}
\end{lstlisting}

\paragraph{Key fields.}
\begin{itemize}[leftmargin=2em]
  \item \texttt{goal\_position\_nominal}: 3D world coordinates of the target object center.
  \item \texttt{sector\_order}: Active sector indices. With 12 total sectors and \texttt{sector\_skip=1}, the even-indexed sectors $\{0, 2, 4, 6, 8, 10\}$ are used (60\textdegree\ spacing).
  \item \texttt{ranges}: Distance ranges for near and far viewpoints (meters from the target).
  \item \texttt{mask\_meets\_threshold}: Whether the target's segmentation mask exceeds the category-specific pixel threshold; \texttt{false} marks a trap view.
  \item \texttt{mask\_bbox\_xyxy}: Bounding box of the target mask in $[x_{\min}, y_{\min}, x_{\max}, y_{\max}]$ format, used as the ground-truth bounding box.
\end{itemize}

\subsection{Evaluation and Training Index (\texttt{.jsonl})}
\label{app:index_format}

The evaluation and training indices share the same JSONL format. Each line defines one episode--pair combination. Listing~\ref{lst:index_examples} shows one entry per pair type.

\begin{lstlisting}[
  caption={Index entries for each pair type. Fields are identical across evaluation and training indices; training indices omit \texttt{depth\_dir}.},
  label={lst:index_examples},
  basicstyle=\ttfamily\scriptsize,
  frame=single,
  breaklines=true,
  numbers=left,
  numberstyle=\tiny\color{gray},
  xleftmargin=2em
]
// Positive pair: target_object_id == query_object_id (same instance)
{
  "episode_path": "pin_capture/val/00813-svBbv1Pavdk/43",
  "scene": "00813-svBbv1Pavdk",
  "episode": "43",
  "meta_path": "pin_capture/val/.../43/meta.json",
  "rgb_dir": "pin_capture/val/.../43/rgb",
  "target_object_id": "8dac2731fff9431399c01ee114e5e002",
  "target_object_category": "laptop",
  "valid_start_sectors": [0, 6, 8, 10],
  "navigable_sectors": [0, 2, 4, 6, 8, 10],
  "n_navigable": 6,
  "n_mask_visible": 4,
  "query_object_id": "8dac2731fff9431399c01ee114e5e002",
  "query_object_category": "laptop",
  "label": 1,
  "pair_type": "positive"
}

// Neg_same pair: different instance, same category
{
  "episode_path": "pin_capture/val/00821-eF36g7L6Z9M/61",
  ...
  "target_object_id": "13615076f9bc40e3b2e6384fd61b12d3",
  "target_object_category": "keys",
  ...
  "query_object_id": "e258f14fb7874f299ebef320f7aee4ee",
  "query_object_category": "keys",
  "label": 0,
  "pair_type": "neg_same"
}

// Neg_diff pair: different instance, different category
{
  "episode_path": "pin_capture/val/00844-q5QZSEeHe5g/6",
  ...
  "target_object_id": "6495988c6c044c76a2fc9f9278543c16",
  "target_object_category": "laptop",
  ...
  "query_object_id": "a3275806b3ed4715b83b2343b93ff8ba",
  "query_object_category": "toy",
  "label": 0,
  "pair_type": "neg_diff"
}
\end{lstlisting}

\paragraph{Key fields.}
\begin{itemize}[leftmargin=2em]
  \item \texttt{target\_object\_id}: The object physically present in the scene (navigated around during capture).
  \item \texttt{query\_object\_id}: The object whose descriptions are used as the verification query. For positive pairs, these are identical; for negative pairs, they differ.
  \item \texttt{valid\_start\_sectors}: Sectors where \texttt{mask\_meets\_threshold=true}, used as valid starting positions (ensures the agent begins with a view of the target).
  \item \texttt{navigable\_sectors}: All sectors with at least one navigable viewpoint. Note that navigable sectors $\supseteq$ valid start sectors: a sector can be navigable but have no visible target (trap view).
  \item \texttt{label}: 1 for positive (match), 0 for negative (no match).
\end{itemize}

\subsection{SFT Training Data}
\label{app:sft_format}

Each SFT training sample is a complete multi-turn conversation with dual-image input. We ship two CoT-label variants: \emph{Generic} (\texttt{sft\_data\_v2.jsonl}) phrases rejection with a coarse placeholder (``expected X but observed something different''), while \emph{Specific} (\texttt{sft\_data\_v3.jsonl}) fills in the observed attribute value (``target should be X, but this object appears Y''). Both share the same conversation skeleton, system prompt, positive-pair CoT, and navigation logic; only the negative-pair rejection phrasing differs. Listing~\ref{lst:sft_example} shows a Generic-CoT sample; Listing~\ref{lst:sft_example_v3} shows the same sample under Specific-CoT.

\begin{lstlisting}[
  caption={Generic-CoT SFT training sample. The assistant response contains template-filled CoT reasoning in \texttt{<think>} and the decision in \texttt{<answer>}. This example shows a neg\_same pair where one attribute is flagged as mismatched.},
  label={lst:sft_example},
  basicstyle=\ttfamily\scriptsize,
  frame=single,
  breaklines=true,
  numbers=left,
  numberstyle=\tiny\color{gray},
  xleftmargin=2em
]
{
  "messages": [
    {
      "role": "system",
      "content": "You are an embodied agent navigating around
        an object to verify whether it matches a target
        description. Each step you see two images: a full
        scene view and a close-up of the detected object.
        You must verify object attributes and decide your
        next action."
    },
    {
      "role": "user",
      "content": "<image><image>\nYou are an embodied agent
        verifying whether a detected object matches a target
        description.\n\nTarget: \"a pair of black and red
        boots with black and white laces and a galaxy print;
        ...\"\nCategory (must match): shoes\nCurrent sector:
        front (0 deg)\nVisited sectors: [front]\nRemaining
        budget: 5 steps\nAvailable sectors: [front-left,
        back-left, back, back-right, front-right]\n\n
        From the scene image (Image 1) and the object
        close-up (Image 2):\n1. Does this object match the
        target description? Check each attribute.\n2. Your
        action: STOP (if confident) or MOVE <sector>
        (if need more views)"
    },
    {
      "role": "assistant",
      "content": "<think>\n[Navigation] Available unvisited
        sectors: front-left, back-left, back, back-right,
        front-right.\n\n[Verification]\n- color: Unsure --
        partially visible, cannot confirm \"black and red\".
        \n- part_attribute_1: No -- expected \"black and
        white\" but observed something different.\n-
        pattern_1: Unsure -- partially visible, cannot
        confirm \"galaxy print\".\n...\n\nAttribute
        \"part_attribute_1\" does not match. This is not
        the target.\n\n[Decision] Found mismatching
        attribute. Confident this is NOT the target.
        Stopping with No.\n</think>\n\n<answer>\n
        verification: No\naction: STOP\n</answer>"
    }
  ],
  "images": [
    "pin_capture/train_sft/.../rgb/rgb_s6_far.png",
    "train_sft/crops/5c4681...6_neg_same.jpg"
  ]
}
\end{lstlisting}

\begin{lstlisting}[
  caption={Specific-CoT SFT training sample (assistant block only; the rest of the conversation is identical to the Generic-CoT example above). The negative-pair rejection now fills in the observed attribute value rather than using the generic ``observed something different'' placeholder.},
  label={lst:sft_example_v3},
  basicstyle=\ttfamily\scriptsize,
  frame=single,
  breaklines=true,
  numbers=left,
  numberstyle=\tiny\color{gray},
  xleftmargin=2em
]
{
  "role": "assistant",
  "content": "<think>\n[Navigation] Available unvisited
    sectors: front-left, back-left, back, back-right,
    front-right.\n\n[Verification]\n- color: Unsure --
    partially visible, cannot confirm \"black and red\".
    \n- part_attribute_1: No -- target should be \"black
    and white\", but this object appears \"grey\".\n-
    pattern_1: Unsure -- partially visible, cannot
    confirm \"galaxy print\".\n...\n\nAttribute
    \"part_attribute_1\" does not match. This is not
    the target.\n\n[Decision] Found mismatching
    attribute. Confident this is NOT the target.
    Stopping with No.\n</think>\n\n<answer>\n
    verification: No\naction: STOP\n</answer>"
}
\end{lstlisting}

\paragraph{Format notes.}
\begin{itemize}[leftmargin=2em]
  \item \texttt{<image>} tags in the user message are placeholders that the training framework (ms-swift) replaces with the corresponding entries from the \texttt{images} array. Image~1 is the scene RGB; Image~2 is the GT bbox crop.
  \item The \texttt{<think>} block follows the \texttt{[Navigation]{-}{-}[Verification]{-}{-}[Decision]} template. Attribute judgments (Yes/No/Unsure) are templated from the ground-truth pair information and per-step visibility metadata supplied by the benchmark, not from visual analysis of the images.
  \item For multi-step trajectories, previous observations are appended to the user prompt (e.g., ``Step 1 (back): Verification=Unsure, Action=MOVE front'').
  \item Directions are always \emph{relative} to the agent's current position: the current sector is always rendered as ``front (0\textdegree)''.
\end{itemize}

\subsection{RL Training Data}
\label{app:rl_format}

RL training samples share the same conversation format as SFT but omit the assistant response (the model generates its own during training). An additional \texttt{solution} field encodes the ground truth for reward computation. The RL pool ships a single version (\texttt{rl\_data\_v2.jsonl}); since RL prompts contain no CoT target, the Generic / Specific distinction does not apply at this stage: the same RL prompts are used regardless of which SFT initialization the agent starts from. Listing~\ref{lst:rl_example} shows the structure.

\begin{lstlisting}[
  caption={Generic-CoT RL training sample. The \texttt{solution} field is a JSON string used by the reward function to evaluate the model's generated responses. Note: no assistant message is provided.},
  label={lst:rl_example},
  basicstyle=\ttfamily\scriptsize,
  frame=single,
  breaklines=true,
  numbers=left,
  numberstyle=\tiny\color{gray},
  xleftmargin=2em
]
{
  "messages": [
    {
      "role": "system",
      "content": "You are an embodied agent navigating around
        an object to verify whether it matches a target
        description. ..."
    },
    {
      "role": "user",
      "content": "<image><image>\n...\nTarget: \"a white mug
        with green text and printed teapots; a white mug
        with decorations referring to tea, like a teapot
        with green tea written on it and another with earl
        grey written on it, ...\"\nCategory (must match):
        mug\nCurrent sector: front (0 deg)\nVisited
        sectors: [back, front]\nRemaining budget: 4 steps
        \nAvailable sectors: [front-left, back-left,
        back-right, front-right]\n\n...\n\nPrevious
        observations:\n- Step 1 (back): Verification=Unsure,
        Action=MOVE front"
    }
  ],
  "images": [
    "pin_capture/train_rl/.../rgb/rgb_s8_far.png",
    "train_rl/crops_rl/rl_8cdaa8...8_positive.jpg"
  ],
  "solution": "{
    \"visible\": [\"back\", \"front\"],
    \"navigable\": [\"back\", \"back-left\", \"back-right\",
                    \"front\", \"front-left\", \"front-right\"],
    \"best_sectors\": [],
    \"label\": \"Yes\",
    \"action\": \"STOP\",
    \"pair_type\": \"positive\"
  }"
}
\end{lstlisting}

\paragraph{Solution fields.}
\begin{itemize}[leftmargin=2em]
  \item \texttt{visible}: Sectors where the target is visible (\texttt{mask\_meets\_threshold=true}). Used to assign the 4-tier action reward: navigating to a visible sector earns 0.7, while a non-visible but navigable sector earns only 0.3.
  \item \texttt{navigable}: All physically reachable sectors.
  \item \texttt{best\_sectors}: FPS-ranked unvisited visible sectors, maximizing angular distance from visited positions; concretely, we keep all visible candidates whose minimum angular distance to the visited set is within 30\textdegree\ of the best candidate (matching \texttt{training/prepare\_rl\_data.py}).
  \item \texttt{label}: Ground-truth verification answer (Yes/No/Unsure).
  \item \texttt{pair\_type}: Used to determine expected behavior (e.g., neg\_diff should STOP immediately).
\end{itemize}

\subsection{DPO Training Data (\texttt{dpo\_data\_v3.jsonl})}
\label{app:dpo_format}

DPO operates only on the Specific-CoT variant: the per-attribute granularity is needed to construct meaningful preference pairs, whereas Generic-CoT's coarse rejection phrasing collapses too many distinct errors onto the same wording. Each sample contains \texttt{messages} (with the \emph{chosen} assistant response as the last message), \texttt{images}, and a \texttt{rejected\_response} field with a perturbed assistant response. Listing~\ref{lst:dpo_example} shows the chosen / rejected pair (other fields identical to the SFT format).

\begin{lstlisting}[
  caption={DPO training sample (Specific-CoT). The chosen response is the original SFT target; the rejected response is a perturbation that flips verification, drops attribute reasoning, or substitutes navigation direction.},
  label={lst:dpo_example},
  basicstyle=\ttfamily\scriptsize,
  frame=single,
  breaklines=true,
  numbers=left,
  numberstyle=\tiny\color{gray},
  xleftmargin=2em
]
// chosen (last message in `messages`):
"<think>\n[Navigation] Available unvisited sectors: ...
   No further exploration needed.\n\n[Verification]\n
- color: Yes -- \"black\" is consistent with what is
   observed.\n- part_attribute_1: Yes -- \"three control
   buttons\" is consistent with what is observed.\n
- part_attribute_2: Yes -- \"three control buttons\" is
   consistent with what is observed.\n- part_attribute_3:
   Yes -- \"metal button\" is consistent with what is
   observed.\n\nAll attributes match the target
   description.\n\n[Decision] All attributes verified as
   matching. Confident this is the target. Stopping
   with Yes.\n</think>\n\n<answer>\nverification: Yes\n
   action: STOP\n</answer>"

// rejected_response (perturbed: flipped verification +
// attribute reasoning weakened to Unsure):
"<think>\n[Navigation] Available unvisited sectors: ...
   \n\n[Verification]\n- color: No -- expected \"black\"
   but observed something different.\n- part_attribute_1:
   Unsure -- cannot fully confirm \"three control buttons\"
   from this view.\n- part_attribute_2: Unsure -- cannot
   fully confirm \"three control buttons\".\n- part_
   attribute_3: Unsure -- cannot fully confirm \"metal
   button\".\n\nFound mismatching attribute. This is not
   the target.\n\n[Decision] Found mismatching attribute.
   Confident this is NOT the target. Stopping with No.\n
   </think>\n\n<answer>\nverification: No\naction: STOP\n
   </answer>"
\end{lstlisting}

\subsection{Object Descriptions (\texttt{object\_descriptions.json})}
\label{app:descriptions}

Each object is associated with three human-written descriptions that characterize its visual appearance. Listing~\ref{lst:descriptions} shows examples for two objects.

\begin{lstlisting}[
  caption={Object descriptions. Each object has three independently written descriptions that highlight different visual attributes. These serve as the verification query.},
  label={lst:descriptions},
  basicstyle=\ttfamily\scriptsize,
  frame=single,
  breaklines=true,
  numbers=left,
  numberstyle=\tiny\color{gray},
  xleftmargin=2em
]
{
  "2ad86321197a49feb54b7726743d7fd0": [
    "a brown leather backpack with gold details
     and two golden buckles",
    "a leather backpack with golden finishes",
    "a chocolate bag"
  ],
  "0907ce5d49bf41e58082f8a51114e79b": [
    "a yellow glovo backpack with white details
     and black shoulder straps",
    "a yellow thermal backpack for food delivery",
    "a yellow cooler with a green logo on it"
  ]
}
\end{lstlisting}

Description quality varies within an object, ranging from specific (``brown leather backpack with gold details'') to vague (``a chocolate bag''). \bench{} preserves this variation rather than normalizing it.

\section{Efficiency--Accuracy Trade-off}
\label{app:eff_acc}

\begin{figure}[ht]
  \centering
  \includegraphics[width=0.85\linewidth]{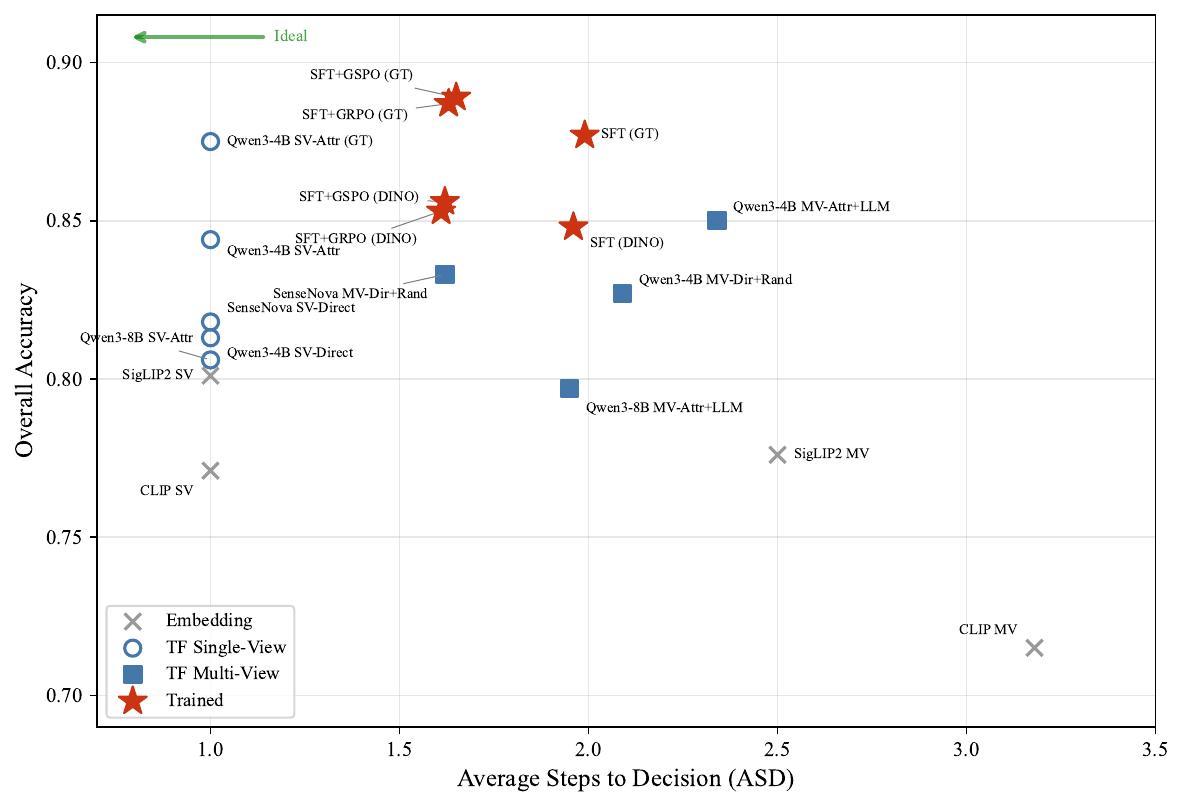}
  \caption{\textbf{Efficiency--accuracy trade-off across all methods.} Trained agents (stars) appear closer to the upper-left region; embedding baselines (crosses) lag. The ideal operating point is the upper-left corner.}
  \label{fig:eff_acc}
\end{figure}

Figure~\ref{fig:eff_acc} visualizes the accuracy--efficiency trade-off across all methods (\S\ref{sec:trained_results}). Trained agents (stars) appear closer to the upper-left region; embedding baselines (crosses) lag. The ideal operating point is the upper-left corner.

\section{Per-Category Breakdown}
\label{app:per_category}

\begin{figure}[ht]
  \centering
  \includegraphics[width=\linewidth]{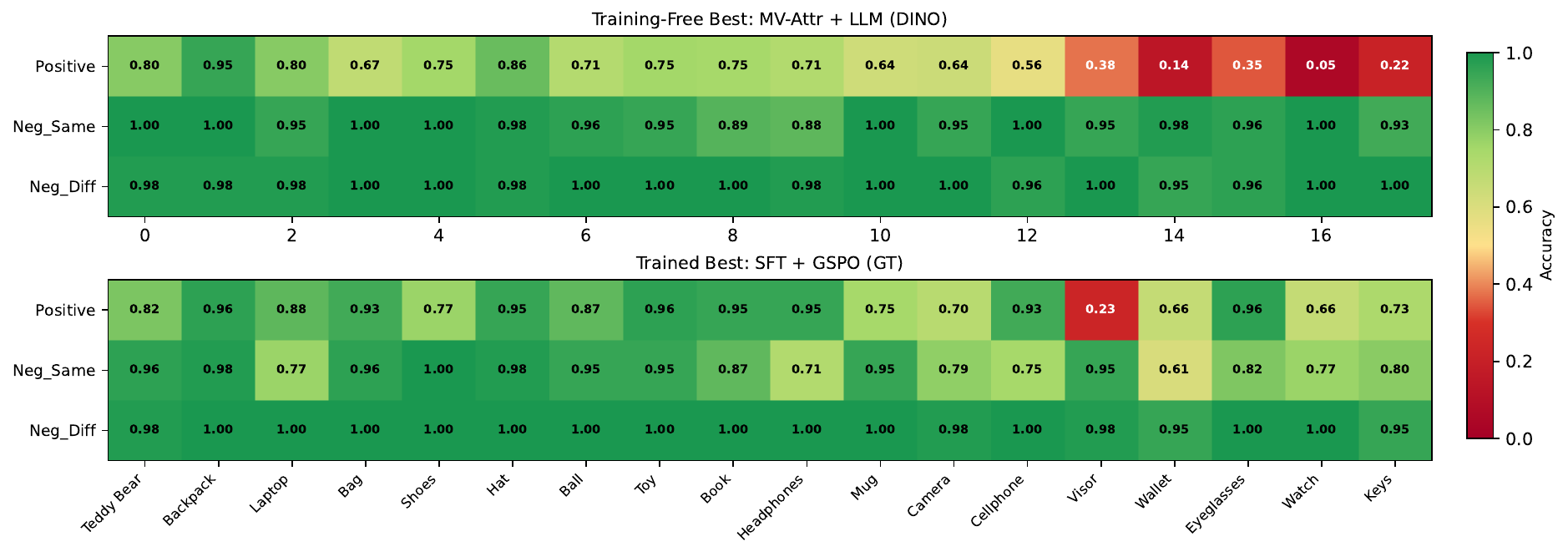}
  \caption{\textbf{Per-category accuracy by pair type} for the main reported training-free agent (MV-Attr+LLM) and the main reported trained agent (SFT+GSPO). Categories are shown in a fixed benchmark order. Failure modes are largely complementary across paradigms.}
  \label{fig:per_cat_heatmap}
\end{figure}

Figure~\ref{fig:per_cat_heatmap} visualizes per-category accuracy by pair type for the main reported training-free agent (MV-Attr+LLM) and the main reported trained agent (SFT+GSPO); Table~\ref{tab:per_category_full} lists the same numbers (DINO detection, Qwen3-VL-4B). Categories are shown in the benchmark category order used by the capture-statistics figures. Failure modes are largely complementary across paradigms: the training-free pipeline is stronger on neg\_same rejection (especially on small objects), while the trained agent is stronger on positive confirmation.

\begin{table}[ht]
\centering
\caption{Per-category accuracy (\%) by pair type for MV-Attr+LLM and SFT+GSPO (DINO detection, Qwen3-VL-4B). Categories are shown in a fixed benchmark order.}
\label{tab:per_category_full}
\setlength{\tabcolsep}{4pt}
\footnotesize
\begin{tabular}{l ccc c ccc}
\toprule
& \multicolumn{3}{c}{\textbf{MV-Attr+LLM}} & & \multicolumn{3}{c}{\textbf{SFT+GSPO}} \\
\cmidrule{2-4} \cmidrule{6-8}
\textbf{Category} & \textbf{Pos} & \textbf{N\_S} & \textbf{N\_D} & & \textbf{Pos} & \textbf{N\_S} & \textbf{N\_D} \\
\midrule
backpack     & 94.55 & 100.00 & 100.00 &  & 96.36 & 98.18 & 100.00 \\
bag          & 67.27 & 100.00 & 96.72  &  & 85.45 & 100.00 & 100.00 \\
teddy bear   & 80.36 & 98.21  & 100.00 &  & 87.50 & 96.43  & 100.00 \\
toy          & 75.00 & 94.64  & 98.21  &  & 92.86 & 91.07  & 100.00 \\
hat          & 85.71 & 98.21  & 94.44  &  & 85.71 & 94.64  & 98.15  \\
laptop       & 80.36 & 94.64  & 98.46  &  & 91.07 & 76.79  & 100.00 \\
ball         & 70.91 & 96.36  & 100.00 &  & 74.55 & 96.36  & 100.00 \\
shoes        & 75.00 & 100.00 & 100.00 &  & 67.86 & 100.00 & 100.00 \\
book         & 74.51 & 89.09  & 95.35  &  & 87.27 & 90.91  & 95.35  \\
headphones   & 71.43 & 87.50  & 98.39  &  & 91.07 & 75.00  & 98.39  \\
mug          & 63.64 & 100.00 & 100.00 &  & 69.09 & 96.36  & 100.00 \\
cellphone    & 56.36 & 100.00 & 100.00 &  & 85.45 & 67.27  & 96.08  \\
eyeglasses   & 34.55 & 96.36  & 100.00 &  & 81.82 & 74.55  & 100.00 \\
camera       & 64.29 & 94.64  & 98.21  &  & 46.43 & 85.71  & 98.21  \\
keys         & 21.82 & 92.73  & 100.00 &  & 61.82 & 70.91  & 98.18  \\
visor        & 37.50 & 94.64  & 98.21  &  & 23.21 & 87.50  & 100.00 \\
watch        &  5.36 & 100.00 & 100.00 &  & 60.71 & 48.21  & 91.94  \\
wallet       & 14.29 & 98.18  & 100.00 &  & 53.57 & 60.71  & 95.56  \\
\midrule
\textbf{Overall} & 59.60 & 96.50 & 98.80 &  & 74.50 & 83.90 & 98.50 \\
\bottomrule
\end{tabular}
\end{table}

\FloatBarrier

\section{Diagnostic Analyses}
\label{app:diagnostics}

This appendix provides four diagnostic tables that support the main-paper observations about NBV similarity (\S\ref{sec:tf_results}) and detection quality (\S\ref{sec:detection_bottleneck}).

\subsection{NBV Strategies: Navigation Failures and Efficiency}
\label{app:nbv_diag}

Table~\ref{tab:nav_failure} reports total navigation failures over the 3{,}000-episode evaluation set, decomposed into unreachable sectors and trap views. Random and LLM-NBV land within 6 failures of each other; FPS suffers $\sim$160 more unreachable hits because it always targets the geometrically farthest sector (often blocked by walls/furniture). Trap-view failures (709--736) are similar across strategies, consistent with many of them being driven by scene-specific occlusion rather than the high-level planner alone.

\begin{table}[ht]
\centering
\caption{Navigation failure analysis (DINO mode, Qwen3-VL-4B, attribute decomposition, multi-view adaptive stopping over 3{,}000 episodes).}
\label{tab:nav_failure}
\small
\begin{tabular}{l c c c}
\toprule
\textbf{NBV} & \textbf{Unreachable} & \textbf{Trap views} & \textbf{Total} \\
\midrule
Random      & 1{,}064 & 729 & 1{,}793 \\
Angular FPS & 1{,}222 & 736 & 1{,}958 \\
LLM-NBV     & 1{,}078 & 709 & 1{,}787 \\
\bottomrule
\end{tabular}
\end{table}

Table~\ref{tab:nbv_efficiency} contrasts NBV efficiency between training-free and trained agents. Training-free agents only trigger multi-view exploration on 38.4\% of episodes and, when they do, average $\sim$3.5 MOVE commands per multi-step episode. SFT triggers multi-step exploration more often (62.1\%) but uses fewer MOVE commands per multi-step episode (1.55). The post-SFT alignment agents (GRPO/GSPO) navigate on 58.6--59.0\% of episodes and average only 1.04--1.05 MOVE commands per multi-step episode: they almost always commit after a single additional view. This accounts for much of the ASD gap (1.61--1.62 vs.\ 2.34--2.36) despite a higher multi-step rate.

\begin{table}[ht]
\centering
\caption{NBV efficiency. ``Multi-step \%'' = episodes requiring more than one view; ``Total NBV'' = MOVE commands across all 3{,}000 episodes; ``NBV/multi-step'' = average MOVEs per multi-step episode.}
\label{tab:nbv_efficiency}
\small
\begin{tabular}{l c c c c}
\toprule
\textbf{Method} & \textbf{Multi-step \%} & \textbf{ASD} & \textbf{Total NBV} & \textbf{NBV/multi-step} \\
\midrule
Random      & 38.4 & 2.34 & 4{,}010 & 3.48 \\
Angular FPS & 38.4 & 2.36 & 4{,}069 & 3.53 \\
LLM-NBV     & 38.4 & 2.34 & 4{,}012 & 3.48 \\
\midrule
SFT         & 62.1 & 1.96 & 2{,}879 & 1.55 \\
SFT+GRPO    & 58.6 & 1.61 & 1{,}829 & 1.04 \\
SFT+GSPO    & 59.0 & 1.62 & 1{,}854 & 1.05 \\
\bottomrule
\end{tabular}
\end{table}

\subsection{Positive-Pair Failure Root-Cause Decomposition}
\label{app:pos_failure}

Across all training-free configurations, positive-pair confirmation lags negative-pair rejection by 15--50\,pp. To approximate failure sources, we classify every failed positive episode by inspecting the final attribute tracker state: an attribute marked Contradictory means the model observed it but judged it a mismatch (\textbf{rejection bias}); an attribute marked Missing means it was never successfully observed (\textbf{trap / observation gap}). Table~\ref{tab:pos_failure} shows the breakdown across five configurations.

\begin{table}[ht]
\centering
\caption{Root-cause decomposition of positive-pair failures by dominant cause (final tracker state). $N_\text{fail}$ = number of failed positive episodes (out of 1{,}000). Mixed: both rejection and observation gap present.}
\label{tab:pos_failure}
\small
\setlength{\tabcolsep}{4pt}
\begin{tabular}{l c c c c c}
\toprule
\textbf{Configuration} & \textbf{Pos Acc} & $N_\text{fail}$ & \textbf{Rejection} & \textbf{Trap/Obs} & \textbf{Mixed} \\
\midrule
SV-Attr (GT)        & 75.8\% & 242 & 75.6\% & 24.4\% &  0.0\% \\
MV-Attr+LLM (GT)    & 72.8\% & 272 & 52.6\% & 33.5\% & 13.9\% \\
SV-Attr (DINO)      & 65.2\% & 348 & 93.7\% &  6.3\% &  0.0\% \\
MV-Attr+Rand (DINO) & 59.2\% & 408 & 82.6\% &  8.3\% &  9.1\% \\
MV-Attr+LLM (DINO)  & 59.6\% & 404 & 83.2\% &  8.2\% &  8.7\% \\
\bottomrule
\end{tabular}
\end{table}

Three observations follow. \emph{(i) MLLM rejection bias is prominent} (53--94\% of failures): even with GT bounding boxes, three quarters of single-view positive failures stem from the model marking observed attributes as Contradictory rather than from missing observations. The effect is amplified under DINO (93.7\% in SV-DINO vs.\ 75.6\% in SV-GT), likely because imprecise crops introduce visual noise that further triggers the model's conservative rejection tendency. The fact that SFT improves Pos from $0.146$ to $0.759$ suggests that calibration or decision-boundary effects contribute beyond pure observation gaps. \emph{(ii) Trap views and observation gaps remain non-trivial in multi-view GT} (33.5\% of failures): aggregating more views can expose more trap-view crops, which feed spurious Contradictory votes into the tracker. \emph{(iii) A simple description-length proxy appears secondary}: failed-episode descriptions average 52.0 characters vs.\ 57.5 for successful ones (SV-GT), a modest gap.

\subsection{Per-Category Detection Quality and the Pos Gap}
\label{app:det_quality}

The GT vs.\ DINO Pos accuracy gap (+10.6\,pp on SV-Attr) is not uniform across categories. Table~\ref{tab:det_quality} shows that small, flat objects with low average detection confidence ($\bar{c}$) suffer the largest gap: \textit{wallet} ($\bar{c}{=}0.272$, $\Delta\text{Pos}{=}+37.5$\,pp) and \textit{watch} ($\bar{c}{=}0.292$, $\Delta\text{Pos}{=}+48.2$\,pp). Larger objects (\textit{teddy bear} $\bar{c}{=}0.612$, \textit{backpack} $\bar{c}{=}0.577$) have higher DINO confidence and slightly higher DINO than GT Pos accuracy in this slice. The Pearson correlation between $\bar{c}$ and Pos accuracy across the 6 representative categories is $r{=}0.932$ (95\% CI $[0.49,\,0.99]$, $n{=}6$, $p{<}0.01$); this is a small-$n$ correlation but is internally consistent with the rejection-bias decomposition above.

\begin{table}[ht]
\centering
\caption{Per-category Grounding DINO confidence vs.\ Pos accuracy gap (MV-Attr+FPS, DINO mode), sorted by $\bar{c}$. ``Low'' = fraction of detections with confidence $<$0.35. $\Delta$Pos = GT $-$ DINO.}
\label{tab:det_quality}
\small
\setlength{\tabcolsep}{4pt}
\begin{tabular}{l c c c c c c}
\toprule
\textbf{Category} & $\bar{c}$ & \textbf{Low\%} & \textbf{Acc} & \textbf{DINO Pos} & \textbf{GT Pos} & $\Delta$\textbf{Pos} \\
\midrule
wallet      & 0.272 & 67\% & 68.5\% & 12.5\% & 50.0\% & $+$37.5 \\
watch       & 0.292 & 63\% & 68.5\% &  5.4\% & 53.6\% & $+$48.2 \\
keys        & 0.412 & 20\% & 72.1\% & 23.6\% & 47.3\% & $+$23.7 \\
camera      & 0.420 & 28\% & 85.7\% & 62.5\% & 89.3\% & $+$26.8 \\
\midrule
backpack    & 0.577 & 13\% & 96.4\% & 90.9\% & 87.3\% & $-$3.6 \\
teddy bear  & 0.612 & 14\% & 93.5\% & 82.1\% & 80.4\% & $-$1.7 \\
\midrule
\multicolumn{4}{l}{\emph{Pearson $r(\bar{c},\,\text{overall})$}} & \multicolumn{3}{c}{0.934} \\
\multicolumn{4}{l}{\emph{Pearson $r(\bar{c},\,\text{Pos})$}}     & \multicolumn{3}{c}{0.932} \\
\bottomrule
\end{tabular}
\end{table}

Together, Tables~\ref{tab:pos_failure} and~\ref{tab:det_quality} support the main-paper observation that detection quality is a major measured bottleneck for positive-pair verification, and identify object categories that may benefit from a stronger open-vocabulary detector.

\section{Qualitative Case Studies}
\label{app:case_studies}

\begin{figure*}[t]
\centering
\includegraphics[width=\linewidth]{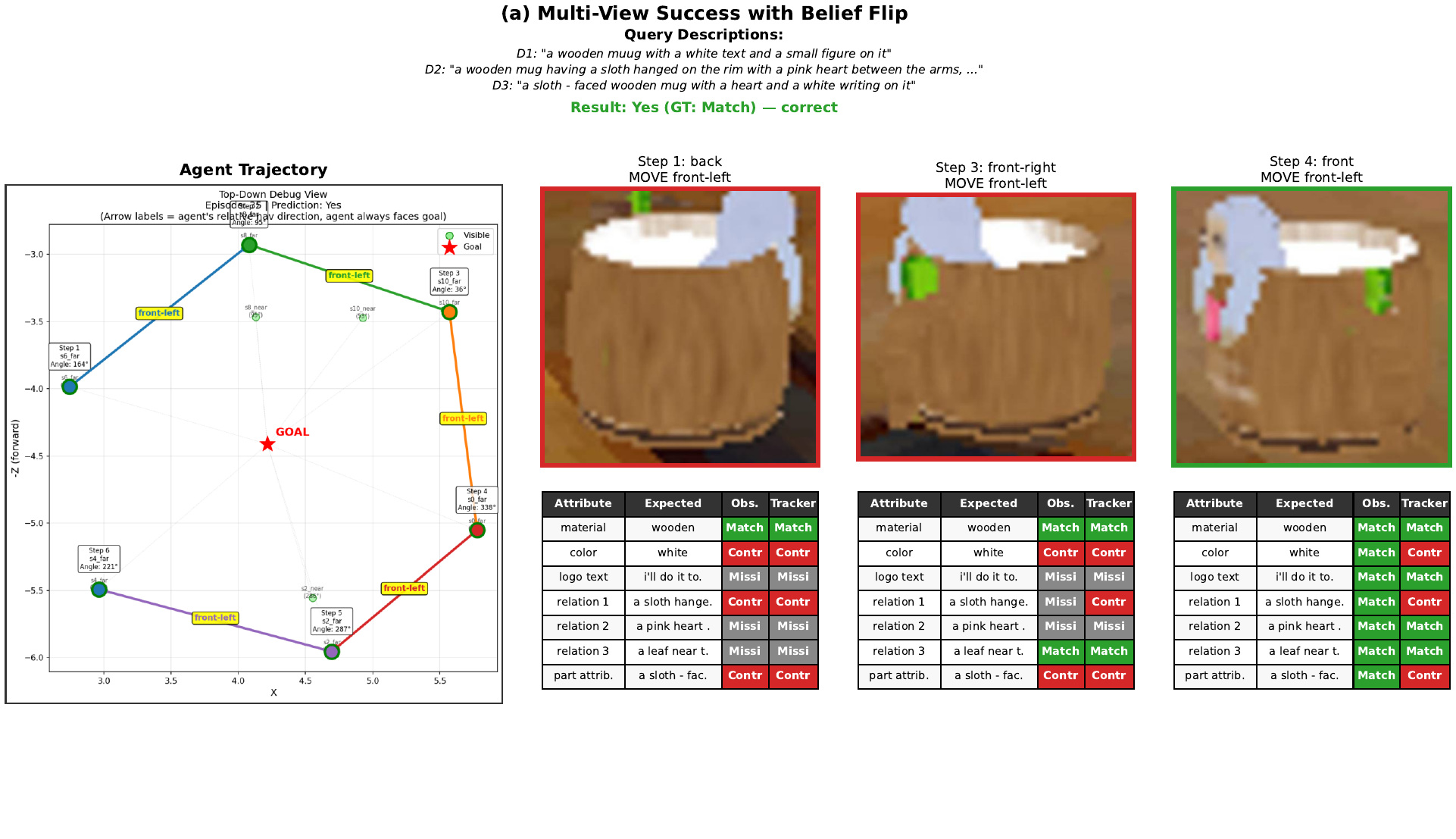}
\caption{\textbf{(a) Multi-view positive verification with a belief flip} (mug, positive pair, MV-Attr+LLM, Qwen3-VL-4B + DINO). At Step~1 (back view) only the \texttt{material} attribute is matched; the sloth figure, the ``I'll do it tomorrow'' text, and the heart are not yet visible, yielding a Mismatch prediction. At Step~3 (front-right) a leaf near the handle is verified but most attributes remain missing. At Step~4 (front), the agent finally observes the text, sloth, and heart, flipping the prediction to Match. The typo ``muug'' in description D1 is inherited from the underlying PInNED descriptions and preserved as-is.}
\label{fig:case_success}
\end{figure*}

\begin{figure*}[t]
\centering
\includegraphics[width=\linewidth]{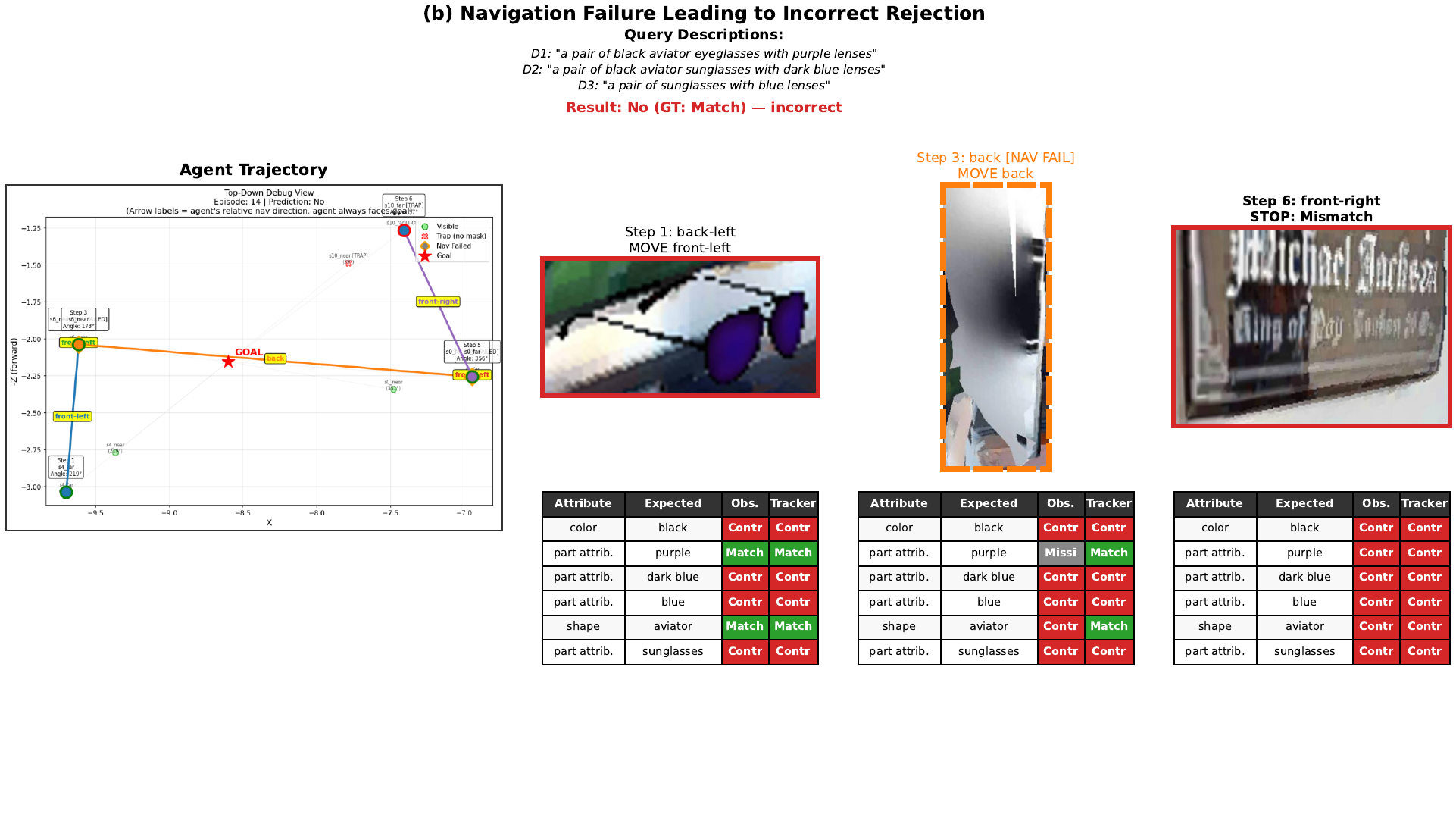}
\caption{\textbf{(b) Navigation failure leading to incorrect rejection} (eyeglasses, positive pair). At Step~1 (back-left) the purple lens colour is matched but most attributes are contradicted under an oblique view. At Step~3 (back) the attempted \texttt{MOVE front-left} fails (orange dashed border) and the agent ends up with an uninformative crop. By Step~6 (front-right) all six attributes are contradicted in the tracker, leading to an incorrect Mismatch despite the object being the correct target. This is a concrete failure mode that the trap-view / unreachable-sector annotations expose.}
\label{fig:case_navfail}
\end{figure*}

\begin{figure*}[t]
\centering
\includegraphics[width=\linewidth]{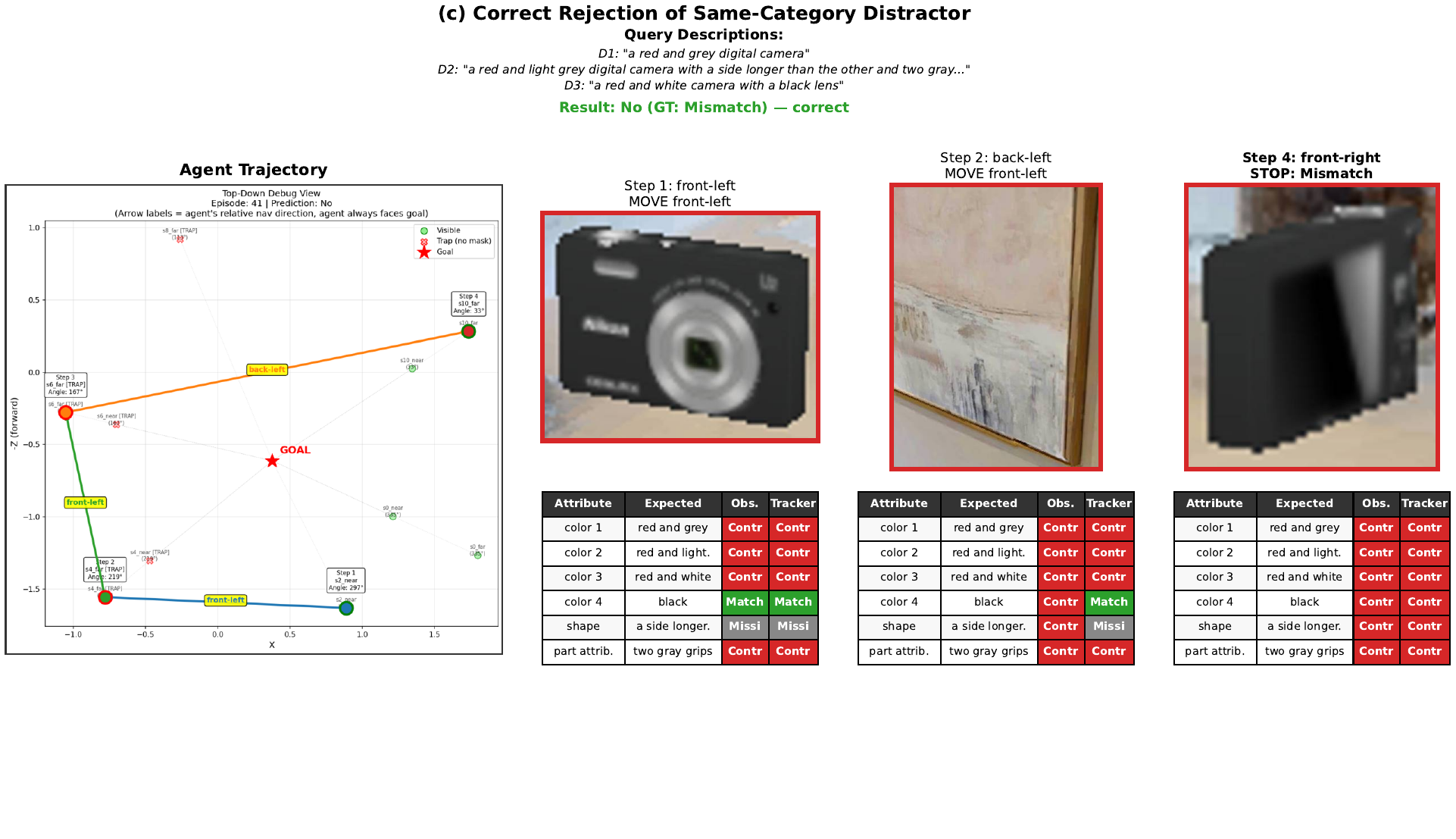}
\caption{\textbf{(c) Correct rejection of a same-category distractor} (camera, neg\_same pair). The query describes ``a red and grey digital camera,'' but the scene contains a black Nikon camera. All three colour attributes (red/grey, red/light grey, red/white) are consistently contradicted across views. The agent explores three viewpoints over four steps and accumulates mismatch evidence on six attributes before issuing a confident rejection at Step~4.}
\label{fig:case_rejection}
\end{figure*}

\begin{figure*}[t]
\centering
\includegraphics[width=\linewidth]{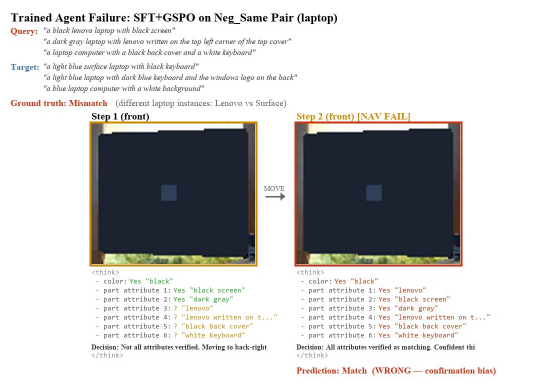}
\caption{\textbf{(d) Trained agent confirmation bias} (laptop, neg\_same pair, SFT+GSPO). The query describes a ``black Lenovo laptop,'' but the target is a different light-blue Surface laptop. The \texttt{<think>} blocks reveal the agent's reasoning: at Step~1 four attributes remain Unsure and the agent decides to explore further. Navigation, however, fails (the agent stays at the same sector), and at Step~2 it confirms \emph{all} seven attributes as matching (including ``lenovo'' and ``white keyboard''), despite viewing essentially the same image. This illustrates the confirmation-bias trade-off observed after SFT: positive accuracy improves by +14.9\,pp at the cost of reduced neg\_same rejection.}
\label{fig:case_trained_failure}
\end{figure*}

To make the verification procedure concrete, Figures~\ref{fig:case_success}--\ref{fig:case_trained_failure} walk through four representative episodes. Each figure shows the agent's top-down trajectory (left) alongside the object crop and per-attribute verification table at selected steps (right). Cells are coloured by per-view outcome (Matched~/~Contradictory~/~Missing); the rightmost column shows the accumulated belief after the visibility-weighted reconciliation of \S\ref{sec:agent}. Together they illustrate (i)~how positive verification builds up via attribute confirmation across views, (ii)~how navigation failures starve the tracker, and (iii)~how the trained end-to-end agent's confirmation bias surfaces concretely.

\paragraph{Label conventions} Throughout these figures, sector labels at each step (e.g., \emph{back-left}, \emph{front-right}) denote the agent's \emph{absolute} position around the goal in a goal-anchored frame, while \texttt{MOVE} directions are \emph{relative} to the agent's current heading (the agent always sees itself as facing ``front'').

\FloatBarrier


\section{Complete Results}
\label{app:full_results}

Full per-configuration results on the 3{,}000-episode evaluation set (1{,}000 per pair type), Grounding DINO detection unless stated otherwise. ASD = Average Steps to Decision. NavFail = navigation failure rate (\%).

\subsection{Embedding-Based Baselines}

Table~\ref{tab:app_embedding} reports all embedding-based configurations. ``Merged'' denotes the variant where three descriptions are concatenated into a single query.

\small
\setlength{\tabcolsep}{4pt}
\begin{longtable}{l l c c c c c}
\caption[Complete embedding baseline results]{Complete results for CLIP and SigLIP2 baselines (Grounding DINO, 3{,}000 episodes).} \label{tab:app_embedding} \\
\toprule
\textbf{Model} & \textbf{Configuration} & \textbf{Overall} & \textbf{Pos} & \textbf{Neg\_S} & \textbf{Neg\_D} & \textbf{ASD} \\
\midrule
\endfirsthead
\toprule
\textbf{Model} & \textbf{Configuration} & \textbf{Overall} & \textbf{Pos} & \textbf{Neg\_S} & \textbf{Neg\_D} & \textbf{ASD} \\
\midrule
\endhead
\midrule
\multicolumn{7}{r}{\footnotesize\textit{continued on next page}} \\
\endfoot
\bottomrule
\endlastfoot
\multicolumn{7}{l}{\textit{CLIP}} \\
& SV & 0.767 & 0.341 & 0.960 & 1.000 & 1.00 \\
& SV-Merged & 0.771 & 0.349 & 0.965 & 1.000 & 1.00 \\
& MV-FPS & 0.717 & 0.163 & 0.987 & 1.000 & 3.28 \\
& MV-FPS-Merged & 0.715 & 0.151 & 0.993 & 1.000 & 3.18 \\
& MV-Random & 0.716 & 0.161 & 0.987 & 1.000 & 3.19 \\
& MV-Random-Merged & 0.716 & 0.156 & 0.993 & 1.000 & 3.08 \\
\midrule
\multicolumn{7}{l}{\textit{SigLIP2}} \\
& SV & 0.755 & 0.267 & 0.999 & 1.000 & 1.00 \\
& SV-Merged & 0.801 & 0.429 & 0.974 & 1.000 & 1.00 \\
& MV-FPS & 0.723 & 0.170 & 1.000 & 1.000 & 2.61 \\
& MV-FPS-Merged & 0.768 & 0.315 & 0.990 & 1.000 & 2.66 \\
& MV-Random & 0.725 & 0.175 & 1.000 & 1.000 & 2.44 \\
& MV-Random-Merged & 0.776 & 0.343 & 0.985 & 1.000 & 2.50 \\
\end{longtable}

SigLIP2 single-view collapses toward a near-degenerate ``always reject'' policy (Pos$=$0.267, Neg\_S$=$0.999), consistent with the calibration analysis in \S\ref{sec:cross_model}.

\needspace{8\baselineskip}
\subsection{MLLM-Based Methods: Qwen3-VL-4B}

Table~\ref{tab:app_qwen4b} reports all Qwen3-VL-4B configurations under Grounding DINO detection. Table~\ref{tab:app_qwen4b_gt} reports GT bounding box results for the subset of configurations where GT was evaluated.

\begin{longtable}{l l c c c c c}
\caption[Complete Qwen3-VL-4B results (DINO)]{Complete results for Qwen3-VL-4B (Grounding DINO, 3{,}000 episodes).} \label{tab:app_qwen4b} \\
\toprule
\textbf{Mode} & \textbf{NBV} & \textbf{Overall} & \textbf{Pos} & \textbf{Neg\_S} & \textbf{Neg\_D} & \textbf{ASD} \\
\midrule
\endfirsthead
\toprule
\textbf{Mode} & \textbf{NBV} & \textbf{Overall} & \textbf{Pos} & \textbf{Neg\_S} & \textbf{Neg\_D} & \textbf{ASD} \\
\midrule
\endhead
\midrule
\multicolumn{7}{r}{\footnotesize\textit{continued on next page}} \\
\endfoot
\bottomrule
\endlastfoot
\multicolumn{7}{l}{\textit{Single-View}} \\
SV-Attr & --- & 0.844 & 0.652 & 0.918 & 0.961 & 1.00 \\
SV-Direct & --- & 0.813 & 0.457 & 0.984 & 0.999 & 1.00 \\
SV-Merged & --- & 0.815 & 0.468 & 0.982 & 0.996 & 1.00 \\
\midrule
\multicolumn{7}{l}{\textit{Multi-View + Attribute Decomposition}} \\
MV-Attr & Random & 0.849 & 0.592 & 0.965 & 0.989 & 2.34 \\
MV-Attr & FPS & 0.848 & 0.592 & 0.964 & 0.989 & 2.36 \\
MV-Attr & LLM & 0.850 & 0.596 & 0.965 & 0.988 & 2.34 \\
\midrule
\multicolumn{7}{l}{\textit{Multi-View + Direct Query}} \\
MV-Direct & Random & 0.827 & 0.492 & 0.991 & 0.999 & 2.09 \\
MV-Direct & FPS & 0.826 & 0.490 & 0.990 & 0.999 & 2.13 \\
MV-Direct & LLM & 0.825 & 0.486 & 0.990 & 0.999 & 2.08 \\
\end{longtable}

\begin{table}[htbp]
\centering
\caption[Qwen3-VL-4B GT bounding box results]{Qwen3-VL-4B results with GT bounding boxes (3{,}000 episodes).}
\label{tab:app_qwen4b_gt}
\small
\begin{tabular}{l l c c c c c}
\toprule
\textbf{Mode} & \textbf{NBV} & \textbf{Overall} & \textbf{Pos} & \textbf{Neg\_S} & \textbf{Neg\_D} & \textbf{ASD} \\
\midrule
SV-Attr & --- & 0.875 & 0.758 & 0.910 & 0.957 & 1.00 \\
SV-Direct & --- & 0.823 & 0.486 & 0.987 & 0.997 & 1.00 \\
MV-Attr & LLM & 0.884 & 0.728 & 0.936 & 0.987 & 2.81 \\
\bottomrule
\end{tabular}
\end{table}

\needspace{8\baselineskip}
\subsection{MLLM-Based Methods: Qwen3-VL-8B}

\noindent
\begin{longtable}{l l c c c c c}
\caption[Complete Qwen3-VL-8B results]{Complete results for Qwen3-VL-8B (Grounding DINO, 3{,}000 episodes).} \label{tab:app_qwen8b} \\
\toprule
\textbf{Mode} & \textbf{NBV} & \textbf{Overall} & \textbf{Pos} & \textbf{Neg\_S} & \textbf{Neg\_D} & \textbf{ASD} \\
\midrule
\endfirsthead
\toprule
\textbf{Mode} & \textbf{NBV} & \textbf{Overall} & \textbf{Pos} & \textbf{Neg\_S} & \textbf{Neg\_D} & \textbf{ASD} \\
\midrule
\endhead
\midrule
\multicolumn{7}{r}{\footnotesize\textit{continued on next page}} \\
\endfoot
\bottomrule
\endlastfoot
\multicolumn{7}{l}{\textit{Single-View}} \\
SV-Attr & --- & 0.806 & 0.446 & 0.973 & 0.998 & 1.00 \\
SV-Direct & --- & 0.749 & 0.249 & 0.998 & 1.000 & 1.00 \\
SV-Merged & --- & 0.776 & 0.331 & 0.996 & 1.000 & 1.00 \\
\midrule
\multicolumn{7}{l}{\textit{Multi-View + Attribute Decomposition}} \\
MV-Attr & Random & 0.796 & 0.410 & 0.979 & 1.000 & 1.95 \\
MV-Attr & FPS & 0.796 & 0.409 & 0.979 & 1.000 & 1.97 \\
MV-Attr & LLM & 0.797 & 0.412 & 0.979 & 1.000 & 1.94 \\
\midrule
\multicolumn{7}{l}{\textit{Multi-View + Direct Query}} \\
MV-Direct & Random & 0.764 & 0.294 & 0.997 & 1.000 & 1.93 \\
MV-Direct & FPS & 0.762 & 0.288 & 0.997 & 1.000 & 1.94 \\
MV-Direct & LLM & 0.763 & 0.292 & 0.997 & 1.000 & 1.91 \\
\end{longtable}

\needspace{8\baselineskip}
\subsection{MLLM-Based Methods: SenseNova-SI-1.2-InternVL3-8B}

\begin{longtable}{l l c c c c c}
\caption[Complete SenseNova results]{Complete results for SenseNova-SI-1.2-InternVL3-8B (Grounding DINO, 3{,}000 episodes).} \label{tab:app_sensenova} \\
\toprule
\textbf{Mode} & \textbf{NBV} & \textbf{Overall} & \textbf{Pos} & \textbf{Neg\_S} & \textbf{Neg\_D} & \textbf{ASD} \\
\midrule
\endfirsthead
\toprule
\textbf{Mode} & \textbf{NBV} & \textbf{Overall} & \textbf{Pos} & \textbf{Neg\_S} & \textbf{Neg\_D} & \textbf{ASD} \\
\midrule
\endhead
\midrule
\multicolumn{7}{r}{\footnotesize\textit{continued on next page}} \\
\endfoot
\bottomrule
\endlastfoot
\multicolumn{7}{l}{\textit{Single-View}} \\
SV-Attr & --- & 0.676 & 0.851 & 0.484 & 0.694 & 1.00 \\
SV-Direct & --- & 0.818 & 0.671 & 0.844 & 0.938 & 1.00 \\
SV-Merged & --- & 0.766 & 0.792 & 0.715 & 0.791 & 1.00 \\
\midrule
\multicolumn{7}{l}{\textit{Multi-View + Attribute Decomposition}} \\
MV-Attr & Random & 0.702 & 0.856 & 0.516 & 0.735 & 2.51 \\
MV-Attr & FPS & 0.703 & 0.857 & 0.527 & 0.724 & 2.56 \\
MV-Attr & LLM & 0.705 & 0.854 & 0.526 & 0.734 & 2.52 \\
\midrule
\multicolumn{7}{l}{\textit{Multi-View + Direct Query}} \\
MV-Direct & Random & 0.833 & 0.658 & 0.884 & 0.958 & 1.62 \\
MV-Direct & FPS & 0.832 & 0.661 & 0.878 & 0.956 & 1.63 \\
MV-Direct & LLM & 0.833 & 0.661 & 0.880 & 0.958 & 1.63 \\
\end{longtable}

\needspace{8\baselineskip}
\subsection{Trained Agents}

Table~\ref{tab:app_trained_all} reports all trained-agent configurations under both Grounding DINO and GT detection modes. We compare two CoT-label variants used during SFT: Generic (coarse rejection phrasing) and Specific (concrete rejection with target/observed attribute pair). NavFail is the navigation failure rate (\%).

\small
\setlength{\tabcolsep}{3pt}
\begin{longtable}{l l l c c c c c c}
\caption{Complete trained-agent results (Qwen3-VL-4B + LoRA, 3{,}000 episodes).} \label{tab:app_trained_all} \\
\toprule
\textbf{CoT} & \textbf{Method} & \textbf{Det.} & \textbf{Overall} & \textbf{Pos} & \textbf{Neg\_S} & \textbf{Neg\_D} & \textbf{ASD} & \textbf{NavFail} \\
\midrule
\endfirsthead
\toprule
\textbf{CoT} & \textbf{Method} & \textbf{Det.} & \textbf{Overall} & \textbf{Pos} & \textbf{Neg\_S} & \textbf{Neg\_D} & \textbf{ASD} & \textbf{NavFail} \\
\midrule
\endhead
\midrule
\multicolumn{9}{r}{\footnotesize\textit{continued on next page}} \\
\endfoot
\bottomrule
\endlastfoot
\multicolumn{9}{l}{\textit{Base (no fine-tuning)}} \\
--- & Base & DINO & 0.706 & 0.146 & 0.973 & 0.999 & 1.74 & 16.3 \\
--- & Base & GT & 0.710 & 0.161 & 0.970 & 0.999 & 1.69 & 15.6 \\
\midrule
\multicolumn{9}{l}{\textit{Generic CoT}} \\
Generic & SFT & DINO & 0.848 & 0.759 & 0.814 & 0.971 & 1.96 & 18.2 \\
Generic & SFT & GT & 0.877 & 0.828 & 0.821 & 0.983 & 1.99 & 19.3 \\
Generic & SFT+GRPO & DINO & 0.853 & 0.736 & 0.838 & 0.985 & 1.61 & 9.0 \\
Generic & SFT+GRPO & GT & 0.887 & 0.806 & 0.863 & 0.993 & 1.63 & 10.1 \\
Generic & SFT+GSPO & DINO & 0.856 & 0.745 & 0.839 & 0.985 & 1.62 & 9.1 \\
Generic & SFT+GSPO & GT & 0.889 & 0.813 & 0.864 & 0.991 & 1.65 & 9.9 \\
\midrule
\multicolumn{9}{l}{\textit{Specific CoT}} \\
Specific & SFT & DINO & 0.858 & 0.697 & 0.885 & 0.991 & 1.92 & 19.0 \\
Specific & SFT & GT & 0.884 & 0.761 & 0.897 & 0.995 & 1.96 & 20.3 \\
Specific & SFT+DPO-200 & DINO & 0.859 & 0.700 & 0.886 & 0.991 & 1.91 & 19.3 \\
Specific & SFT+DPO-200 & GT & 0.881 & 0.756 & 0.893 & 0.994 & 1.96 & 20.7 \\
Specific & SFT+DPO-400 & DINO & 0.860 & 0.665 & 0.921 & 0.994 & 1.91 & 19.2 \\
Specific & SFT+DPO-400 & GT & 0.884 & 0.729 & 0.927 & 0.997 & 1.96 & 20.4 \\
Specific & SFT+GRPO & DINO & 0.855 & 0.793 & 0.792 & 0.980 & 1.60 & 10.4 \\
Specific & SFT+GRPO & GT & 0.884 & 0.847 & 0.813 & 0.991 & 1.63 & 11.2 \\
Specific & SFT+GSPO & DINO & 0.851 & 0.796 & 0.781 & 0.977 & 1.56 & 8.9 \\
Specific & SFT+GSPO & GT & 0.889 & 0.813 & 0.864 & 0.991 & 1.65 & 9.9 \\
\end{longtable}

\needspace{8\baselineskip}
\subsection{Evaluation Running Time}

Table~\ref{tab:runtime_eval} reports the evaluation (inference) running time for all 3{,}000 episodes on the validation set. All times assume pre-computed caches (category, merged-description, and attribute caches) are loaded. All experiments use NVIDIA RTX 3090 GPUs. Wall time is the actual elapsed time; GPU time $=$ wall time $\times$ number of GPUs.

\begin{longtable}{llccc}
\caption[Evaluation running time]{Evaluation running time for all 3{,}000 episodes.} \label{tab:runtime_eval} \\
\toprule
\textbf{Model} & \textbf{Configuration} & \textbf{GPUs} & \textbf{Wall Time} & \textbf{GPU Time} \\
\midrule
\endfirsthead
\toprule
\textbf{Model} & \textbf{Configuration} & \textbf{GPUs} & \textbf{Wall Time} & \textbf{GPU Time} \\
\midrule
\endhead
\midrule
\multicolumn{5}{r}{\footnotesize\textit{continued on next page}} \\
\endfoot
\bottomrule
\endlastfoot
\multicolumn{5}{l}{\textit{CLIP}} \\
& SV & 1 & 10\,min & 10\,min \\
& SV-Merged & 1 & 10\,min & 10\,min \\
& MV-FPS & 1 & 38\,min & 38\,min \\
& MV-FPS-Merged & 1 & 37\,min & 37\,min \\
& MV-Random & 1 & 38\,min & 38\,min \\
& MV-Random-Merged & 1 & 36\,min & 36\,min \\
\midrule
\multicolumn{5}{l}{\textit{SigLIP2}} \\
& SV & 1 & 13\,min & 13\,min \\
& SV-Merged & 1 & 14\,min & 14\,min \\
& MV-FPS & 1 & 35\,min & 35\,min \\
& MV-FPS-Merged & 1 & 36\,min & 36\,min \\
& MV-Random & 1 & 34\,min & 34\,min \\
& MV-Random-Merged & 1 & 34\,min & 34\,min \\
\midrule
\multicolumn{5}{l}{\textit{Qwen3-VL-4B (Training-Free, DINO)}} \\
& SV-Attr & 8 & 1h\,18m & 10h\,24m \\
& SV-Direct & 8 & 52\,min & 6h\,56m \\
& SV-Merged & 8 & 30\,min & 4h\,00m \\
& MV-Attr+FPS & 8 & 3h\,14m & 25h\,52m \\
& MV-Attr+Random & 8 & 3h\,14m & 25h\,52m \\
& MV-Attr+LLM & 8 & 3h\,53m & 31h\,04m \\
& MV-Direct+FPS & 8 & 1h\,55m & 15h\,20m \\
& MV-Direct+Random & 8 & 1h\,53m & 15h\,04m \\
& MV-Direct+LLM & 8 & 2h\,21m & 18h\,48m \\
\midrule
\multicolumn{5}{l}{\textit{Qwen3-VL-4B (Training-Free, GT)}} \\
& SV-Attr (GT) & 8 & 1h\,27m & 11h\,36m \\
& SV-Direct (GT) & 8 & 56\,min & 7h\,28m \\
& MV-Attr+LLM (GT) & 8 & 5h\,22m & 42h\,56m \\
\midrule
\multicolumn{5}{l}{\textit{Qwen3-VL-8B (Training-Free, DINO)}} \\
& SV-Attr & 8 & 1h\,08m & 9h\,04m \\
& SV-Direct & 8 & 50\,min & 6h\,40m \\
& SV-Merged & 8 & 30\,min & 4h\,00m \\
& MV-Attr+FPS & 8 & 2h\,09m & 17h\,12m \\
& MV-Attr+Random & 8 & 2h\,07m & 16h\,56m \\
& MV-Attr+LLM & 8 & 3h\,11m & 25h\,28m \\
& MV-Direct+FPS & 8 & 1h\,27m & 11h\,36m \\
& MV-Direct+Random & 8 & 1h\,27m & 11h\,36m \\
& MV-Direct+LLM & 8 & 2h\,23m & 19h\,04m \\
\midrule
\multicolumn{5}{l}{\textit{SenseNova-SI-1.2-InternVL3-8B (Training-Free, DINO)}} \\
& SV-Attr & 8 & 1h\,12m & 9h\,36m \\
& SV-Direct & 8 & 51\,min & 6h\,48m \\
& SV-Merged & 8 & 30\,min & 4h\,00m \\
& MV-Attr+FPS & 8 & 2h\,27m & 19h\,36m \\
& MV-Attr+Random & 8 & 2h\,30m & 20h\,00m \\
& MV-Attr+LLM & 8 & 3h\,38m & 29h\,04m \\
& MV-Direct+FPS & 8 & 1h\,08m & 9h\,04m \\
& MV-Direct+Random & 8 & 1h\,08m & 9h\,04m \\
& MV-Direct+LLM & 8 & 1h\,47m & 14h\,16m \\
\midrule
\multicolumn{5}{l}{\textit{Trained Agents --- Generic-CoT (v2)}} \\
Base (no FT) & DINO & 4 & 1h\,02m & 4h\,08m \\
             & GT   & 4 & 59\,min & 3h\,56m \\
SFT          & DINO & 4 & 2h\,49m & 11h\,16m \\
             & GT   & 4 & 2h\,49m & 11h\,16m \\
SFT+GRPO    & DINO & 4 & 2h\,33m & 10h\,12m \\
             & GT   & 4 & 2h\,31m & 10h\,04m \\
SFT+GSPO    & DINO & 4 & 2h\,36m & 10h\,24m \\
             & GT   & 4 & 2h\,37m & 10h\,28m \\
\midrule
\multicolumn{5}{l}{\textit{Trained Agents --- Specific-CoT (v3)}} \\
SFT          & DINO & 4 & 3h\,16m & 13h\,04m \\
             & GT   & 4 & 3h\,04m & 12h\,16m \\
SFT+DPO-200 & DINO & 4 & 3h\,04m & 12h\,16m \\
             & GT   & 4 & 3h\,05m & 12h\,20m \\
SFT+DPO-400 & DINO & 4 & 3h\,06m & 12h\,24m \\
             & GT   & 4 & 3h\,06m & 12h\,24m \\
SFT+GRPO    & DINO & 4 & 2h\,27m & 9h\,48m \\
             & GT   & 4 & 2h\,28m & 9h\,52m \\
SFT+GSPO    & DINO & 4 & 2h\,34m & 10h\,16m \\
             & GT   & 4 & 2h\,36m & 10h\,24m \\
\end{longtable}

\end{document}